\pdfoutput=1
\documentclass[11pt]{article}

\usepackage{acl}

\usepackage{times}
\usepackage{latexsym}

\usepackage[T1]{fontenc}
\usepackage[utf8]{inputenc}

\usepackage{microtype}

\usepackage{booktabs}
\usepackage{tikz}
\usepackage{gnuplot-lua-tikz}
\usepackage{multirow}
\usepackage{amssymb}
\usepackage{amsmath}

\title{Non-Autoregressive Machine Translation: It's Not as Fast as it Seems}

\author{Jind\v{r}ich Helcl{\normalfont\textsuperscript{1,2}} \and
        Barry Haddow{\normalfont\textsuperscript{1}} \and 
        Alexandra Birch{\normalfont\textsuperscript{1}} \\
    \textsuperscript{1}School of Informatics, University of Edinburgh \\
    \textsuperscript{2}Faculty of Mathematics and Physics, Charles University \\
    \texttt{\{jhelcl,bhaddow,a.birch\}@ed.ac.uk} \\
  }

\begin{document}
\maketitle
\begin{abstract}
  Efficient machine translation models are commercially important as they can
  increase inference speeds, and reduce costs and carbon emissions.
  Recently, there has been much interest in non-autoregressive (NAR) models, 
  which promise faster translation.
  In parallel to the research on NAR models, there have been successful attempts
  to create optimized autoregressive models as part of the WMT shared task on efficient translation.
  In this paper, we point out flaws in the evaluation methodology present in the literature on NAR models and we provide a fair comparison between a state-of-the-art
  NAR model and the autoregressive submissions to the shared task.
  We make the case for consistent evaluation of NAR models, and also for the importance of comparing NAR models with other widely used methods for improving efficiency.
  We run experiments with a connectionist-temporal-classification-based (CTC)
  NAR model implemented in C++ and compare it with AR models using wall clock
  times.
  Our results show that, although NAR models are faster on GPUs, with small batch sizes, they are almost always slower under
  more realistic usage conditions. We call for more realistic and extensive evaluation of NAR models in future work.
\end{abstract}

\section{Introduction}

Non-autoregressive neural machine translation (NAR NMT, or NAT;
\citealp{gu2017nonautoregressive, lee-etal-2018-deterministic}) is an emerging subfield of
NMT which focuses on increasing the translation speed by changing the
model architecture.

The defining feature of non-autoregressive models is the conditional
independence assumption on the output probability distributions; this is in contrast to autoregressive models, where the output distributions are conditioned on the previous outputs. This
conditional independence allows one to decode the target tokens in parallel.
This can substantially reduce the decoding time, especially for longer
target sentences.

The decoding speed is assessed by translating a test set and measuring
the overall time the process takes. This may sound simple, but there are various
aspects to be considered that can affect decoding speed, such as
batching, number of hypotheses in
beam search or hardware used (i.e., using CPU or GPU). Decoding speed
evaluation is a challenging task, especially when it comes to
comparability across different approaches. Unlike  translation quality, decoding
speed can be measured exactly. However, also unlike translation quality,
different results are obtained from the same system under different evaluation environments.
The WMT Efficient Translation Shared Task aims to evaluate efficiency research  and encourages the reporting of a range
of speed and translation quality values to better understand the trade-off across different model configurations \citep{heafield-etal-2021-findings}.
In this paper, we follow the emerging best practices developed in the Efficiency Shared Task and directly compare with the submitted systems.

In the development of NAR models, modeling error and its subsequent
negative effect on translation quality remains the biggest issue. Therefore,
the goal of contemporary research is to close the performance gap between the
AR models and their NAR counterparts, while maintaining high decoding speed.
Considering these stated research goals, the evaluation should comprise of
assessing translation quality as well as decoding speed.

Translation quality is usually evaluated by scoring translations of an unseen
test set either using automatic metrics, such as BLEU
\citep{papineni-etal-2002-bleu}, ChrF \citep{popovic-2015-chrf} or COMET
\citep{rei-etal-2020-comet}, or using human evaluation. To prevent  methods
from eventually overfitting to a single test set, new test sets are published
each year as part of the WMT News Translation Shared Task. In contrast,
translation quality evaluation in NAR research is 
measured almost exclusively on the WMT~14 English-German test set, using 
only BLEU scores. Automatic evaluation of translation quality
 remains an open research problem, but current research advises against
 relying on a single metric, and especially against relying on only BLEU
\citep{mathur-etal-2020-tangled,kocmi2021ship}. In our experiments, we follow
the recent best practices by using multiple metrics and recent test sets.

In this paper, we examine the evaluation methodology generally accepted in
literature on NAR methods, and we identify a number of flaws.
First, the results are reported on different hardware architectures, which
makes them incomparable, even when comparing only relative speedups. Second,
most of the methods only report latency (decoding with a single sentence per
batch) using a GPU; we show that this is the only setup favors NAR models. Third, the
reported baseline performance is usually questionable, both in terms of speed
and translation quality. Finally, despite the fact that the main motivation for
using NAR models is the lower time complexity, the findings of the efficiency
task are ignored in most of the NAR papers.

We try to connect the separate worlds of NAR and efficient translation research. We train
non-autoregressive models based on connectionist temporal classification (CTC),
an approach previously shown to be effective
\citep{libovicky-helcl-2018-end, ghazvininejad2020aligned, gu-kong-2021-fully}.  We
employ a number of techniques for improving the translation quality, including
data cleaning and sequence-level knowledge distillation
\citep{kim-rush-2016-sequence}. We evaluate our models following a unified
evaluation methodology: In order to compare the translation quality with the
rest of the NAR literature, we report BLEU scores measured on the WMT~14 test
set, on which we achieve state-of-the-art performance among (both single-step
and iterative) NAR methods; we also evaluate the translation quality and decoding
speed of our models in the same conditions as the efficiency task.

We find that despite achieving very good results among the NAT models on the
WMT~14 test set, our models fall behind in translation quality when measured on
the recent WMT~21 test set using three different automatic evaluation
metrics. Moreover, we show that GPU decoding latency is the only scenario in
which non-autoregressive models outperform autoregressive models.

This paper contributes to the research community in the following aspects:
First, we point out weaknesses in standard evaluation methodology of non-autoregressive models.
Second, we link the worlds of non-autoregressive translation and optimization of autoregressive models to provide a better understanding
of the results achieved in the related work.
%\BH{I'm not sure what `model optimization' means in this context} 
%\JH{provide fair and thorough comparison with state-of-the-art autoregressive translation. }

% -----------------------------------------------------------------------------
\section{Non-Autoregressive NMT}%
\label{sec:nat}
% -----------------------------------------------------------------------------

The current state-of-the-art NMT models are autoregressive -- the output
distributions are conditioned on the previously generated tokens
\citep{bahdanau2016neural, vaswani-etal-2017-attention}. The decoding process
is sequential in its nature, limiting the opportunities for parallelization.

Non-autoregressive models use output distributions which are conditionally
independent of each other, which opens up the possibility of
parallelization. Formally, the probability of a sequence $y$ given the input
$x$ in a non-autoregressive model with parameters $\theta$ is modeled as
\begin{equation}
    p_{\theta}(y|x) = \prod_{y_i \in y} p(y_i | x, \theta).
\end{equation}

Unsurprisingly, the independence assumption in NAR models has a negative impact
on the translation quality. The culprit for this behavior is the
\emph{multimodality problem} -- the inability of the model to differentiate
between different modes of the joint probability distribution over output
sequences inside the distributions corresponding to individual time steps. A
classic example of this issue is the sentence ``Thank you'' with its two
equally probable German translations ``Danke schön'' and ``Vielen Dank''
\citep{gu2017nonautoregressive}. Because of the independence assumption, a non-autoregressive
model cannot assign high probabilities to these two translations without also
allowing for the incorrect sentences ``Vielen schön'' and ``Danke Dank''.

Knowledge distillation \citep{kim-rush-2016-sequence} has been successfully
employed to reduce the negative influence of the multimodality problem in NAR
models \citep{gu2017nonautoregressive, saharia-etal-2020-non}. Synthetic data tends to be less
diverse than authentic texts, therefore the number of equally likely
translation candidates gets smaller \citep{zhou-etal-2020-understanding}.

A number of techniques have been proposed for training NAR models, including iterative methods \citep{lee-etal-2018-deterministic,ghazvininejad-etal-2019-mask},
auxiliary training objectives \citep{wang2019non, qian-etal-2021-glancing}, or latent variables \citep{gu2017nonautoregressive,lee-etal-2018-deterministic,kaiser2018fast}.
In some form, all of the aforementioned approaches use explicit target length estimation, and rely on one-to-one correspondence between the output distributions and the reference sentence.

A group of methods that relax the requirement of the strict one-to-one alignment between the model outputs and the ground-truth target sequence include aligned cross-entropy \citep{ghazvininejad2020aligned} and connectionist temporal classification \citep{libovicky-helcl-2018-end}.

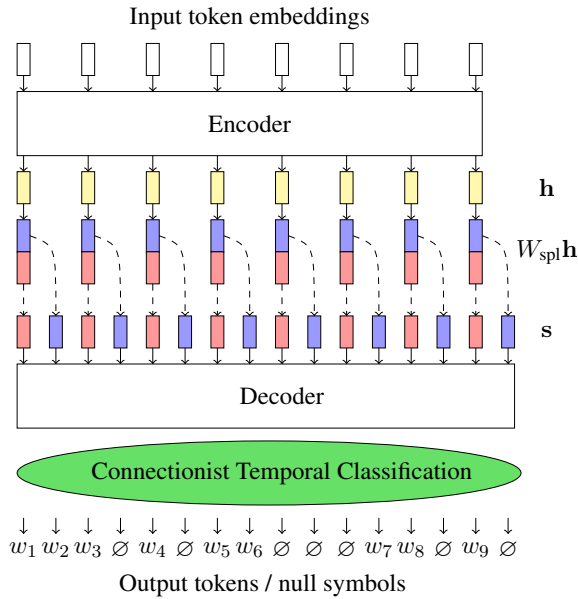
\begin{figure}
  \centering
  \scalebox{0.85}{\def\inputsize{7}

\begin{tikzpicture}[]

\draw (\inputsize / 2 + 0.1, -0.1) node {Input token embeddings};

\foreach \i in {0,...,\inputsize} {
	\draw (\i,-0.5) rectangle (\i+0.2,-1);
    \draw [->] (\i+0.1,-1) -- (\i+0.1, -1.25);
};

\draw (0, -1.25) rectangle (\inputsize + 0.2, -2.25);
\draw (\inputsize / 2 + 0.1, -1.75) node {Encoder};

\foreach \i in {0,...,\inputsize} {
	\draw [->] (\i+0.1,-2.25) -- (\i+0.1, -2.5);
    \draw[fill=yellow!40] (\i,-2.5) rectangle (\i+0.2,-3);

    \draw [->] (\i+0.1,-3) -- (\i+0.1, -3.25);
	\draw[fill=blue!40] (\i,-3.25) rectangle (\i+0.2,-3.75);
	\draw[fill=red!40] (\i,-3.75) rectangle (\i+0.2,-4.25);

    \draw [dashed,->] (\i+0.1,-4.25) -  - (\i+0.1, -4.75);
    \draw [dashed,->] (\i+0.2,-3.5) .. controls (\i + 0.6, -3.65) .. (\i+0.6, -4.75);

	\draw[fill=red!40] (\i,-4.75) rectangle (\i+0.2,-5.25);
	\draw[fill=blue!40] (\i + 0.5,-4.75) rectangle (\i+0.7,-5.25);

    \draw [->] (\i+0.1,-5.25) - - (\i+0.1, -5.5);
    \draw [->] (\i+0.6,-5.25) - - (\i+0.6, -5.5);
};

\draw (\inputsize + 1.2, -2.75) node {$\mathbf{h}$};
\draw (\inputsize + 1.2, -3.75) node {$W_\text{spl}\mathbf{h}$};
\draw (\inputsize + 1.2, -5.00) node {$\mathbf{s}$};

\draw (0, -5.5) rectangle (\inputsize + 0.7, -6.5);
\draw (\inputsize / 2 + 0.5 + 0.1, -6.0) node {Decoder};

\draw [fill=green!80!black!60] (\inputsize / 2 + 0.4,-7.2) circle [x radius=\inputsize / 2 + 0.4, y radius=0.5];
\draw (\inputsize / 2 + 0.6, -7.2) node {Connectionist Temporal Classification};

\foreach \i in {0,...,\inputsize} {
   \draw [->] (\i+0.1,-7.9) - - (\i+0.1, -8.15);
   \draw [->] (\i+0.6,-7.9) - - (\i+0.6, -8.15);
}

\draw  (0+0.1,-8.4) node {$w_1$};
\draw  (0+0.6,-8.4) node {$w_2$};
\draw  (1+0.1,-8.4) node {$w_3$};
\draw  (1+0.6,-8.4) node {$\varnothing$};
\draw  (2+0.1,-8.4) node {$w_4$};
\draw  (2+0.6,-8.4) node {$\varnothing$};
\draw  (3+0.1,-8.4) node {$w_5$};
\draw  (3+0.6,-8.4) node {$w_6$};
\draw  (4+0.1,-8.4) node {$\varnothing$};
\draw  (4+0.6,-8.4) node {$\varnothing$};
\draw  (5+0.1,-8.4) node {$\varnothing$};
\draw  (5+0.6,-8.4) node {$w_7$};
\draw  (6+0.1,-8.4) node {$w_8$};
\draw  (6+0.6,-8.4) node {$\varnothing$};
\draw  (7+0.1,-8.4) node {$w_9$};
\draw  (7+0.6,-8.4) node {$\varnothing$};

\draw (\inputsize / 2 + 0.3, -8.95) node {Output tokens / null symbols};

\end{tikzpicture}}
  \caption{The schema of the CTC-based non-autoregressive architecture. We show
    the original image from \citet{libovicky-helcl-2018-end}.}
  \label{fig:ctc-schema}
\end{figure}

The schema of the CTC-based model, as proposed by \citet{libovicky-helcl-2018-end}, is shown in Figure \ref{fig:ctc-schema}. The model extends
the Transformer architecture \citep{vaswani-etal-2017-attention}. It consists
of an encoder, a state-splitting layer, and a non-autoregressive decoder. The
encoder has the same architecture as in the Transformer model. The
state-splitting layer, applied on the encoder output, linearly projects and
splits each state into $k$ states with the same dimension. The decoder consists
of a stack of Transformer layers. Unlike the Transformer model, the
self-attention in the non-autoregressive decoder does not use the causal mask,
so the model is not prevented from attending to future states. Since the output
length is fixed to $k$-times the length of the source sentence, the model is
permitted to output blank tokens. Different positions of the blank tokens in
the output sequence represent different alignments between the outputs and the
ground-truth sequence.  Connectionist temporal classification
\citep{graves2006connectionist} is a dynamic algorithm that efficiently
computes the standard cross-entropy loss summed over all possible alignments.

We choose the CTC-based
architecture for our models because it has been previously shown to be
effective for NAR NMT \citep{gu-kong-2021-fully,saharia-etal-2020-non} and
performs well in the context of non-autoregressive research. It is also one
of the fastest NAR architectures since it is not iterative.

% \LB{This is good. Could you succinctly describe here why CTC is a good fit for NAR translation? You explained it to me - that the model does not have to commit to exact positions which is important for the multi-modal problem - or something similar. You have to make the case why you chose CTC. Obviously it also performs well in terms of speed and BLEU. Good balance between allowing uncertainty but also efficient - not iterative. Got space to fill here. Maybe even put a small diagram. Don't want to go mad. }

% -----------------------------------------------------------------------------
\section{Evaluation Methodology} %
\label{sec:eval}
% -----------------------------------------------------------------------------

The research goal of the non-autoregressive methods is to improve translation
quality while maintaining the speedup brought by the conditional independence assumption. This means that careful thought should be given to both quantifying the speed gains and the translation quality evaluation. The speed-vs-quality trade-off can be characterized by the Pareto frontier.
In this section we discuss the evaluation from both perspectives. 

\paragraph{Translation Quality.}
In the world of non-autoregressive NMT, the experimental settings are not very diverse. The primary language pair for translation experiments is English-German, sometimes accompanied by English-Romanian to simulate the low-resource scenario. These language pairs, along with the widely used test sets -- WMT~14 \citep{bojar-etal-2014-findings} for En-De and WMT~16 \cite{bojar-etal-2016-findings} for En-Ro --  became the de facto standard benchmark for NAR model evaluation.

A common weakness seen in the literature is the use of weak baseline models.
The base variant of the Transformer model is used almost exclusively \citep{gu2017nonautoregressive, gu-kong-2021-fully, lee-etal-2018-deterministic, ghazvininejad2020aligned, qian-etal-2021-glancing}. We argue that using weaker baselines might lead to overrating the positive effects brought by proposed improvements.
Since the baseline autoregressive models are used to generate the synthetic parallel data for knowledge distillation, the weakness is potentially further amplified in this step.

Evaluation is normally with automatic metrics only, and often only BLEU is reported. In light of recent
research casting further doubt on the reliability of BLEU as a measure of translation quality \cite{kocmi2021ship}, we argue that this is insufficient.

\paragraph{Decoding Speed.}
The current standard in evaluation of NAR models is to measure translation latency using a GPU, i.e., the average time to translate a single sentence without batching. Since the time
depends on the hardware, relative speedup is usually reported along with latency.

This is a reasonable approach but we need to keep in mind the associated difficulties.
First, the results achieved on different hardware architectures are not easily comparable even when considering the relative speedups. We also note that the relative speedup values should always be accompanied by the corresponding decoding times in absolute numbers. Sometimes, this information is missing from the published results \citep{qian-etal-2021-glancing}.

%\JH{Using weak baselines in terms of speed as well.}

We argue that measuring only GPU latency disregards other use-cases.
In the WMT Efficiency Shared Task, the decoding speed is measured in five scenarios. The speed is reported using a GPU with and without batching, using all 36 CPU cores (also, with and without batching), and using a single CPU core without batching. In batched decoding, the shared task participants could choose the optimal batch size.
%
%\LB{This following sentence is maybe not necessary here - it is said in intro and in results} \JH{Until we run out of space, I would keep this here for narrative purposes..}
Our results in Section \ref{sec:results} show that measuring latency is the only one that favors NAR models, and as the batch size increases, AR models quickly reach higher translation speeds.

% % -----------------------------------------------------------------------------
% \section{Optimization Techniques} %
% \label{sec:optim}
% % -----------------------------------------------------------------------------

% \subsection{Knowledge Distillation}

% Besides dealing with the multimodality problem, knowledge distillation is a
% helpful optimization tool as well. In the non-autoregressive research, the
% distilled models have the same size as the teacher models. However, it is also
% possible to train a much smaller model. In autoregressive models, this brings
% improvements in terms of memory consumption and decoding speed, often only for a
% little drop in translation quality.

% \subsection{Shortlisting}

% In an NMT model, the decoder generates the output sequence by projecting the
% hidden state into a vocabulary-sized vector representing the probability
% distribution over the target words. However, given an input batch, a majority
% of the words in vocabulary receive zero probabilities. Narrowing the vocabulary
% to only a small subset of words could therefore speed up the matrix
% multiplication and the subsequent softmax operation at almost no cost in
% modeling performance.

% Given a batch of source sentences, we can use a bilingual lexicon to compose a list of
% most likely candidates to appear on the target side in the batch. This list
% can be used as a shortlist for the vocabulary. \JH{cite}

% -----------------------------------------------------------------------------
\section{Experiments}%
\label{sec:experiments}
% -----------------------------------------------------------------------------
 
We experiment with non-autoregressive models for English-German
translation. We used the data provided by the
WMT~21 News Translation Shared Task organizers
\citep{akhbardeh-etal-2021-findings}.

%\JH{we try and view the results of NAR and efficiency task in shared perspective}
%\LB{This previous sentence would be a good way to introduce the experiments section}
%\JH{i assume you meant the results section -- i moved it there.}

As our baseline model, we use the CTC-based NAR model as described by
\citet{libovicky-helcl-2018-end}. We use stack of 6 encoder and 6 decoder layers, separated by the state splitting layer which extends the state sequence 3 times.

We implement our models\footnote{Our code is publicly available at \url{https://github.com/jindrahelcl/marian-dev}} in the Marian toolkit
\citep{junczys-dowmunt-etal-2018-marian}. For the CTC loss computation, we use
the warp-ctc library \citep{amodei2016deep}.
%\BH{Is the code publicly available? We should mention this.} 

\begin{table}
    \centering
    \begin{tabular}{lrrr}
    \toprule
    Data & Raw size & Cleaned size \\
    \midrule
    Parallel -- clean & 3.9 & 3.1 \\
    Parallel -- noisy & 92.0 & 84.6 \\
    \addlinespace
    Monolingual -- En & 93.1 & 91.0 \\
    Monolingual -- De & 149.9 & 146.2 \\
    \bottomrule
    \end{tabular}
    \caption{The sizes of the parallel and monolingual training datasets (in millions of examples).}
    \label{tab:data}
\end{table}

% -----------------------------------------------------------------------------
\subsection{Teacher Models}
\label{sec:teachers}
% -----------------------------------------------------------------------------

For training our baseline autoregressive models, we closely follow 
the approach of \citet{chen-wmt21-news}. The preparation of the baseline models consists of three phases -- data cleaning, backtranslation, and the training of the final models.

% We train four autoregressive Transformer big models %\BH{Is transformer capitalised? It's not above}\JH{unified to capitalized form}
% in ensembles for generating the knowledge-distilled data. We follow the approach of \citet{chen-wmt21-news}.

We train the teacher models on cleaned parallel corpora and backtranslated
monolingual data. For the parallel data, we used Europarl \citep{koehn-2005-europarl}, the RAPID corpus \citep{rozis-skadins-2017-tilde}, and the News Commentary corpus from OPUS \citep{tiedemann-2012-parallel}.  We consider these three parallel datasets clean. We also use noisier parallel datasets, namely Paracrawl \citep{banon-etal-2020-paracrawl},
Common Crawl\footnote{\url{https://commoncrawl.org/}}, WikiMatrix \citep{schwenk2019wikimatrix}, and
Wikititles\footnote{\url{https://linguatools.org/}}. For backtranslation, we used the monolingual datasets from the News Crawl from the years 2018-2020, in both English and German.

We clean the parallel corpus (i.e. both clean and noisy portions) using rule-based
cleaning\footnote{\url{https://github.com/browsermt/students/blob/master/train-student/clean/clean-corpus.sh}}. Additionally, we exclude sentence pairs with non-latin
characters.
and we apply dual cross-entropy filtering on the noisy part of the parallel data \citep{junczys-dowmunt-2018-dual}. We train
Transformer base models in both directions on the clean portion of the parallel data. Then, we select the best-scoring 75\% of sentence pairs for the final  parallel portion of the training dataset.

For backtranslation \citep{sennrich-etal-2016-improving}, we train four Transformer big models on the cleaned parallel data in both directions. We then use them in an ensemble
to create the synthetic source side for the monolingual corpora. We add a special symbol to the generated sentences to help the models differentiate between synthetic and authentic source language data \citep{caswell-etal-2019-tagged}.

We use hyperparameters of the Transformer big model, i.e. model
dimension 1,024, feed-forward hidden dimension of 4,096, and 16 attention heads. For training, we use the Adam optimizer \citep{kingma2014adam} with $\beta_1$, $\beta_2$ and $\epsilon$ set to 0.9, 0.998 and 10\textsuperscript{-9} respectively. We used the 
inverted square-root learning rate decay with 8,000 steps of linear warm-up and initial learning rate of 10\textsuperscript{-4}.

The teacher models follow the same hyperparameter settings as the models for backtranslation, but are trained with the tagged backtranslations included in the data. As in the previous case, we train four teacher models with different random seeds for ensembling.

Similar to creating the backtranslations, we use the four teacher models in an ensemble to create the knowledge-distilled data \citep{kim-rush-2016-sequence}. We translate the source side of the
parallel data,
as well as the source-language monolingual data.
We do not translate back-translated data. Thus, the source side data
for the student models is authentic, and the target side is synthetic, created
by the teacher models. %\BH{So the student and teacher models actually use different sets of monolingual data (with different sizes), since  one uses the source, and the other the target. I don't think this matters for the claims in this paper.}

% -----------------------------------------------------------------------------
\subsection{Student Models}
% -----------------------------------------------------------------------------

We train five variants of the student models with different hyperparameter settings. 
The ``Large''
model is our baseline model -- the same number of layers as the teacher models,
6 in the encoder, followed by the state splitting layer, and another 6 layers
in the decoder. The ``Base'' model has the same number of layers with reduced
dimension of the embeddings and the feed-forward Transformer sublayer, to match
the Transformer base settings. We also try reducing the numbers of encoder and
decoder layers. We shrink the base model to 3-3 (``Small''), 2-2 (``Micro''), and 1-1
(``Tiny'') architectures.

We run the training of each model for three weeks on four Nvidia Pascal P100 GPUs. 

%\JH{something to add about how did the training go?} \BH{Do you want to say how long it took?}

% -----------------------------------------------------------------------------
\section{Results}%
\label{sec:results}
% -----------------------------------------------------------------------------

In this section, we try to view the results of the NAR and efficiency research in a shared perspective. We evaluate our models and present results in terms of
translation quality and decoding speed. We compare the results to the related
work on both non-autoregressive translation and model optimization.

\paragraph{Translation Quality.} The research on non-autoregressive models uses
the BLEU score \citep{papineni-etal-2002-bleu} measured on the WMT~14 test set
\citep{bojar-etal-2014-findings} as a standard benchmark for evaluating
translation quality. We use Sacrebleu \citep{post-2018-call} as the implementation of the BLEU score metric. %\BH{The signature shows 3 refs -- but WMT 14 only has one}
Using a single test set for the whole volume of research
on this topic may however produce misleading results. To bring the evaluation
up to date with the current state-of-the-art translation systems, we also
evaluate our models using COMET \citep{rei-etal-2020-comet}\footnote{We use the COMET model \texttt{wmt20-comet-da} from version dd2298 (1.0.0.rc9). } and BLEU\footnote{
Signature: \texttt{nrefs:3|bs:1000|seed:12345|\\case:mixed|eff:no|tok:13a|smooth:exp|\\
version:2.0.0}. For WMT~21 De $\rightarrow$ En, only 2 references were used. For WMT~14, we used the  signature with the exception of having only a single reference.} scores on
the recent WMT~21 test set.  The same test set was used in the WMT~21 Efficiency
Task.

Table \ref{tab:wmt14-bleu-scores} shows the BLEU scores of our NAR models on
the WMT~14 test set. We show the results of the five variants of the NAR models
and we include three of the best-performing NAR approaches from the related
work.  We see from the table that using BLEU, the ``Large'' model scores among
the best NAR models on the WMT~14 test set. As the NAR model size decreases, so
does the translation quality, with the notable exception of the
En$\rightarrow$De ``Micro'' model, which outperforms the ``Base'' model
consistently on different test sets.
%\LB{Would be good to hazard a guess as to why?} \BH{It only outperforms on one test set?}

\begin{table}
  \centering
  \begin{tabular}{lcc}
    \toprule
     & En $\rightarrow$ De & De $\rightarrow$ En \\
    \midrule
    %\citep{gu2017nonautoregressive} & 19.17 & 21.47 \\
    \citet{saharia-etal-2020-non} &  28.2 &  31.8 \\
    \citet{gu-kong-2021-fully} & 27.2 & 31.3 \\
    \citet{qian-etal-2021-glancing} & 26.6 & 31.0 \\
    \midrule

    Large & 28.4  & 31.3 \\
    Base  &  23.7  & 30.3 \\
    Small &  23.6  & 29.1 \\
    Micro &  25.0  & 27.5 \\
    Tiny  & 20.3 & 21.7 \\

    \bottomrule
  \end{tabular}

  \caption{The BLEU scores of the NAR models on the WMT~14 test set}%
  \label{tab:wmt14-bleu-scores}
\end{table}

In Table \ref{tab:wmt21}, we report the automatic evaluation results of our AR
and NAR models on the multi-reference WMT~21 test set
\citep{akhbardeh-etal-2021-findings}. We compare our NAR models to the AR large
teacher models from Section \ref{sec:teachers}, an AR base model trained on the
original clean data, and an AR base student model trained on the distilled
data.  Following \citet{heafield-etal-2021-findings}, we use references A, C,
and D for English-German translation.
%, and references A and B for German-English translation.
%\JH{Do we mention de-en? there is nothing to compare to -- the efficiency task was only in en-de.}

We see that there is a considerable difference in the translation quality
between the NAR models and the AR large teacher model.  This difference grows
with beam search and ensembling applied on the AR decoding, techniques not
usually used with NAR models because of the speed cost. We also note that when
we train an AR base model on the distilled data, it outperforms the NAR large
model by a considerable margin.

Another thing we notice is the enormous difference in the COMET scores between
the AR and NAR models. The AR base models achieve comparable BLEU scores to the
NAR large models, but differ substantially in the COMET score. 
From a look at the system outputs, we hypothesize that the NAR systems produce unusual errors which BLEU does not penalise as heavily as COMET. This might suggest that NAR models would rank poorly in
human evaluation relative to their autoregressive counterparts, despite the
reasonable BLEU score values. Another reason might be that the different errors of NAR models are causing a domain mismatch between the COMET training data and the data being evaluated.

% \JH{what do we
%   think about this?} 
% \LB{This is very interesting! Could you eyeball this? Or is it too late?} \BH{We did eyball this a couple of weeks ago. The NAR systems 
% are producing unusual errors (maybe related to CTC?) which BLEU does not penalise heavily, but COMET does}  

\begin{table}
  \centering
  \begin{tabular}{lrr@{}l}
    \toprule
    En $\rightarrow$ De & \multicolumn{1}{c}{COMET} & \multicolumn{2}{c}{BLEU} \\
    \midrule
    AR -- Large & 0.4110 & 50.5 & \small \enspace \textpm 1.3 \\
    \quad + beam & 0.4053 & 50.8 & \small \enspace \textpm 1.3 \\
    \quad + ensemble & 0.4332 & 52.2 & \small \enspace \textpm 1.3 \\
    \addlinespace
    AR -- Base & 0.3881 & 47.9 & \small \enspace \textpm 1.3 \\
    \quad + beam & 0.3873 & 48.0 & \small \enspace \textpm 1.3 \\

    \midrule
    Student AR -- Base   &  0.4550 & 51.6 & \small \enspace \textpm 1.2 \\

    \addlinespace
    NAR models \\
    Large &   0.1485 & 47.8 & \small \enspace \textpm 1.2 \\
    Base  &  -0.0521 & 41.8 & \small \enspace \textpm 1.1 \\
    Small &  -0.0752 & 41.9 & \small \enspace \textpm 1.1 \\
    Micro &  -0.0083 & 43.5 & \small \enspace \textpm 1.1 \\
    Tiny  &  -0.3333 & 34.7 & \small \enspace \textpm 1.0 \\
    \bottomrule

    \end{tabular}
    \caption{Results of quantitative evaluation of English-German translation
      quality using automatic metrics on the multi-reference WMT~21 test
      set. The confidence intervals were computed using Sacrebleu. }
    \label{tab:wmt21}
\end{table}

% \begin{table*}
%   \centering
%   \begin{tabular}{lrrr@{}l}
%     \toprule
%     Model & ChrF & COMET & \multicolumn{2}{c}{BLEU} \\
%     \midrule
%     Single greedy AR \\
%     Large & 59.2 & 0.4110 & 50.5 & \small \enspace \textpm 1.3 \\
%     Base  & 58.1 & 0.3881 & 47.9 & \small \enspace \textpm 1.3 \\

%     \addlinespace
%     Single beam AR \\
%     Large & 58.8 & 0.4053 & 50.8 & \small \enspace \textpm 1.3 \\
%     Base  & 57.9 & 0.3873 & 48.0 & \small \enspace \textpm 1.3 \\

%     \addlinespace
%     Ensemble beam AR \\
%     Large & 59.5 & 0.4332 & 52.2 & \small \enspace \textpm 1.3 \\
%     \midrule
%     Student AR -- Base   & 59.5 &  0.4550 & 51.6 & \small \enspace \textpm 1.2 \\

%     \addlinespace
%     NAR models \\
%     Large & 58.6 &  0.1485 & 47.8 & \small \enspace \textpm 1.2 \\
%     Base  & 56.3 & -0.0521 & 41.8 & \small \enspace \textpm 1.1 \\
%     Small & 56.2 & -0.0752 & 41.9 & \small \enspace \textpm 1.1 \\
%     Micro & 57.3 & -0.0083 & 43.5 & \small \enspace \textpm 1.1 \\
%     Tiny  & 53.6 & -0.3333 & 34.7 & \small \enspace \textpm 1.0 \\
%     \bottomrule

%     \end{tabular}
%     \caption{Results of quantitative evaluation of translation quality using automatic metrics.}
%     \label{tab:wmt19}
% \end{table*}

\paragraph{Decoding speed.} We follow the decoding time evaluation methodology
of the WMT~21 Efficient Translation Shared Task
\citep{heafield-etal-2021-findings}. We recreate the hardware conditions that
were used in the task. For the GPU decoding measurements, we use a single
Nvidia Ampere A100 GPU. The CPU evaluation was performed on a 36-core CPU Intel
Xeon Gold 6354 server from Oracle cloud.
To amortize the various computation overheads, the models submitted to the
shared task are evaluated on a million sentence benchmark dataset.
%\BH{Wasn't this to stop NEU from cheating?}\JH{that's not the official explanation in the paper :)}

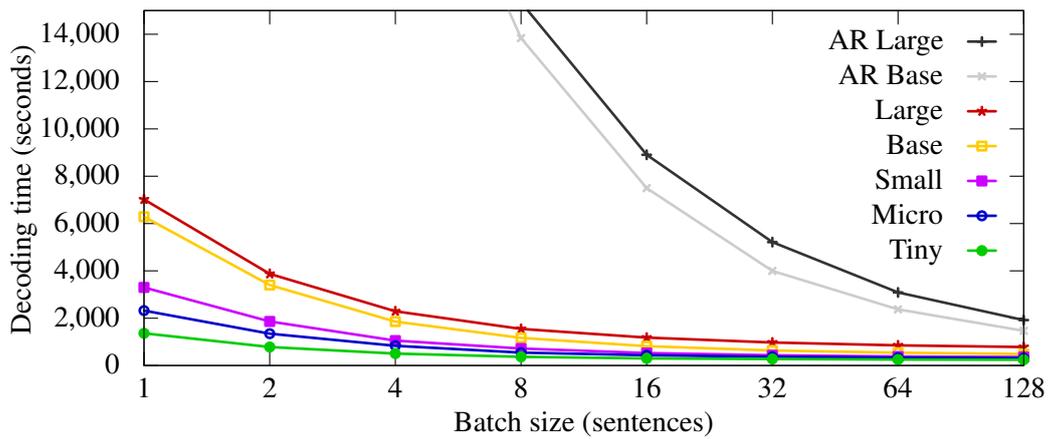
\begin{figure*}
    \centering
    \scalebox{1.0}{\begin{tikzpicture}[gnuplot]
%% generated with GNUPLOT 5.2p8 (Lua 5.3; terminal rev. Nov 2018, script rev. 108)
%% Thu 16 Dec 2021 11:23:05 AM CET
\path (0.000,0.000) rectangle (14.000,6.000);
\gpcolor{color=gp lt color border}
\gpsetlinetype{gp lt border}
\gpsetdashtype{gp dt solid}
\gpsetlinewidth{1.00}
\draw[gp path] (1.872,0.985)--(2.052,0.985);
\draw[gp path] (13.447,0.985)--(13.267,0.985);
\node[gp node right] at (1.688,0.985) {0};
\draw[gp path] (1.872,1.612)--(2.052,1.612);
\draw[gp path] (13.447,1.612)--(13.267,1.612);
\node[gp node right] at (1.688,1.612) {2,000};
\draw[gp path] (1.872,2.240)--(2.052,2.240);
\draw[gp path] (13.447,2.240)--(13.267,2.240);
\node[gp node right] at (1.688,2.240) {4,000};
\draw[gp path] (1.872,2.867)--(2.052,2.867);
\draw[gp path] (13.447,2.867)--(13.267,2.867);
\node[gp node right] at (1.688,2.867) {6,000};
\draw[gp path] (1.872,3.495)--(2.052,3.495);
\draw[gp path] (13.447,3.495)--(13.267,3.495);
\node[gp node right] at (1.688,3.495) {8,000};
\draw[gp path] (1.872,4.122)--(2.052,4.122);
\draw[gp path] (13.447,4.122)--(13.267,4.122);
\node[gp node right] at (1.688,4.122) {10,000};
\draw[gp path] (1.872,4.750)--(2.052,4.750);
\draw[gp path] (13.447,4.750)--(13.267,4.750);
\node[gp node right] at (1.688,4.750) {12,000};
\draw[gp path] (1.872,5.377)--(2.052,5.377);
\draw[gp path] (13.447,5.377)--(13.267,5.377);
\node[gp node right] at (1.688,5.377) {14,000};
\draw[gp path] (1.872,0.985)--(1.872,1.165);
\draw[gp path] (1.872,5.691)--(1.872,5.511);
\node[gp node center] at (1.872,0.677) {1};
\draw[gp path] (3.526,0.985)--(3.526,1.165);
\draw[gp path] (3.526,5.691)--(3.526,5.511);
\node[gp node center] at (3.526,0.677) {2};
\draw[gp path] (5.179,0.985)--(5.179,1.165);
\draw[gp path] (5.179,5.691)--(5.179,5.511);
\node[gp node center] at (5.179,0.677) {4};
\draw[gp path] (6.833,0.985)--(6.833,1.165);
\draw[gp path] (6.833,5.691)--(6.833,5.511);
\node[gp node center] at (6.833,0.677) {8};
\draw[gp path] (8.486,0.985)--(8.486,1.165);
\draw[gp path] (8.486,5.691)--(8.486,5.511);
\node[gp node center] at (8.486,0.677) {16};
\draw[gp path] (10.140,0.985)--(10.140,1.165);
\draw[gp path] (10.140,5.691)--(10.140,5.511);
\node[gp node center] at (10.140,0.677) {32};
\draw[gp path] (11.793,0.985)--(11.793,1.165);
\draw[gp path] (11.793,5.691)--(11.793,5.511);
\node[gp node center] at (11.793,0.677) {64};
\draw[gp path] (13.447,0.985)--(13.447,1.165);
\draw[gp path] (13.447,5.691)--(13.447,5.511);
\node[gp node center] at (13.447,0.677) {128};
\draw[gp path] (1.872,5.691)--(1.872,0.985)--(13.447,0.985)--(13.447,5.691)--cycle;
\node[gp node center,rotate=-270] at (0.292,3.338) {Decoding time (seconds)};
\node[gp node center] at (7.659,0.215) {Batch size (sentences)};
\node[gp node right] at (12.531,5.280) {AR Large};
\gpcolor{rgb color={0.200,0.200,0.200}}
\gpsetlinewidth{2.50}
\draw[gp path] (12.715,5.280)--(13.079,5.280);
\draw[gp path] (6.922,5.691)--(8.486,3.779)--(10.140,2.621)--(11.793,1.954)--(13.447,1.587);
\gpsetpointsize{4.00}
\gppoint{gp mark 1}{(8.486,3.779)}
\gppoint{gp mark 1}{(10.140,2.621)}
\gppoint{gp mark 1}{(11.793,1.954)}
\gppoint{gp mark 1}{(13.447,1.587)}
\gppoint{gp mark 1}{(12.897,5.280)}
\gpcolor{color=gp lt color border}
\node[gp node right] at (12.531,4.818) {AR Base};
\gpcolor{rgb color={0.800,0.800,0.800}}
\draw[gp path] (12.715,4.818)--(13.079,4.818);
\draw[gp path] (6.671,5.691)--(6.833,5.326)--(8.486,3.337)--(10.140,2.239)--(11.793,1.729)%
  --(13.447,1.445);
\gppoint{gp mark 2}{(6.833,5.326)}
\gppoint{gp mark 2}{(8.486,3.337)}
\gppoint{gp mark 2}{(10.140,2.239)}
\gppoint{gp mark 2}{(11.793,1.729)}
\gppoint{gp mark 2}{(13.447,1.445)}
\gppoint{gp mark 2}{(12.897,4.818)}
\gpcolor{color=gp lt color border}
\node[gp node right] at (12.531,4.356) {Large};
\gpcolor{rgb color={0.800,0.000,0.000}}
\draw[gp path] (12.715,4.356)--(13.079,4.356);
\draw[gp path] (1.872,3.187)--(3.526,2.200)--(5.179,1.704)--(6.833,1.470)--(8.486,1.355)%
  --(10.140,1.290)--(11.793,1.252)--(13.447,1.230);
\gppoint{gp mark 3}{(1.872,3.187)}
\gppoint{gp mark 3}{(3.526,2.200)}
\gppoint{gp mark 3}{(5.179,1.704)}
\gppoint{gp mark 3}{(6.833,1.470)}
\gppoint{gp mark 3}{(8.486,1.355)}
\gppoint{gp mark 3}{(10.140,1.290)}
\gppoint{gp mark 3}{(11.793,1.252)}
\gppoint{gp mark 3}{(13.447,1.230)}
\gppoint{gp mark 3}{(12.897,4.356)}
\gpcolor{color=gp lt color border}
\node[gp node right] at (12.531,3.894) {Base};
\gpcolor{rgb color={1.000,0.800,0.000}}
\draw[gp path] (12.715,3.894)--(13.079,3.894);
\draw[gp path] (1.872,2.958)--(3.526,2.052)--(5.179,1.567)--(6.833,1.351)--(8.486,1.241)%
  --(10.140,1.184)--(11.793,1.155)--(13.447,1.137);
\gppoint{gp mark 4}{(1.872,2.958)}
\gppoint{gp mark 4}{(3.526,2.052)}
\gppoint{gp mark 4}{(5.179,1.567)}
\gppoint{gp mark 4}{(6.833,1.351)}
\gppoint{gp mark 4}{(8.486,1.241)}
\gppoint{gp mark 4}{(10.140,1.184)}
\gppoint{gp mark 4}{(11.793,1.155)}
\gppoint{gp mark 4}{(13.447,1.137)}
\gppoint{gp mark 4}{(12.897,3.894)}
\gpcolor{color=gp lt color border}
\node[gp node right] at (12.531,3.432) {Small};
\gpcolor{rgb color={0.800,0.000,1.000}}
\draw[gp path] (12.715,3.432)--(13.079,3.432);
\draw[gp path] (1.872,2.020)--(3.526,1.569)--(5.179,1.315)--(6.833,1.210)--(8.486,1.150)%
  --(10.140,1.121)--(11.793,1.104)--(13.447,1.097);
\gppoint{gp mark 5}{(1.872,2.020)}
\gppoint{gp mark 5}{(3.526,1.569)}
\gppoint{gp mark 5}{(5.179,1.315)}
\gppoint{gp mark 5}{(6.833,1.210)}
\gppoint{gp mark 5}{(8.486,1.150)}
\gppoint{gp mark 5}{(10.140,1.121)}
\gppoint{gp mark 5}{(11.793,1.104)}
\gppoint{gp mark 5}{(13.447,1.097)}
\gppoint{gp mark 5}{(12.897,3.432)}
\gpcolor{color=gp lt color border}
\node[gp node right] at (12.531,2.970) {Micro};
\gpcolor{rgb color={0.000,0.000,0.800}}
\draw[gp path] (12.715,2.970)--(13.079,2.970);
\draw[gp path] (1.872,1.713)--(3.526,1.407)--(5.179,1.246)--(6.833,1.156)--(8.486,1.121)%
  --(10.140,1.100)--(11.793,1.089)--(13.447,1.083);
\gppoint{gp mark 6}{(1.872,1.713)}
\gppoint{gp mark 6}{(3.526,1.407)}
\gppoint{gp mark 6}{(5.179,1.246)}
\gppoint{gp mark 6}{(6.833,1.156)}
\gppoint{gp mark 6}{(8.486,1.121)}
\gppoint{gp mark 6}{(10.140,1.100)}
\gppoint{gp mark 6}{(11.793,1.089)}
\gppoint{gp mark 6}{(13.447,1.083)}
\gppoint{gp mark 6}{(12.897,2.970)}
\gpcolor{color=gp lt color border}
\node[gp node right] at (12.531,2.508) {Tiny};
\gpcolor{rgb color={0.000,0.800,0.000}}
\draw[gp path] (12.715,2.508)--(13.079,2.508);
\draw[gp path] (1.872,1.412)--(3.526,1.230)--(5.179,1.143)--(6.833,1.100)--(8.486,1.079)%
  --(10.140,1.071)--(11.793,1.064)--(13.447,1.061);
\gppoint{gp mark 7}{(1.872,1.412)}
\gppoint{gp mark 7}{(3.526,1.230)}
\gppoint{gp mark 7}{(5.179,1.143)}
\gppoint{gp mark 7}{(6.833,1.100)}
\gppoint{gp mark 7}{(8.486,1.079)}
\gppoint{gp mark 7}{(10.140,1.071)}
\gppoint{gp mark 7}{(11.793,1.064)}
\gppoint{gp mark 7}{(13.447,1.061)}
\gppoint{gp mark 7}{(12.897,2.508)}
\gpcolor{color=gp lt color border}
\gpsetlinewidth{1.00}
\draw[gp path] (1.872,5.691)--(1.872,0.985)--(13.447,0.985)--(13.447,5.691)--cycle;
%% coordinates of the plot area
\gpdefrectangularnode{gp plot 1}{\pgfpoint{1.872cm}{0.985cm}}{\pgfpoint{13.447cm}{5.691cm}}
\end{tikzpicture}
%% gnuplot variables
}
    \caption{The decoding times to translate the efficiency task test set using various batch size settings, computed on a single Nvidia Ampere A100 GPU, i.e. the GPU type used for evaluation in the efficiency task.} 
    \label{fig:time-gpu}
\end{figure*}

\begin{figure*}
    \centering
    \scalebox{1.0}{\begin{tikzpicture}[gnuplot]
%% generated with GNUPLOT 5.2p8 (Lua 5.3; terminal rev. Nov 2018, script rev. 108)
%% Thu 16 Dec 2021 11:23:12 AM CET
\path (0.000,0.000) rectangle (14.000,6.000);
\gpcolor{color=gp lt color border}
\gpsetlinetype{gp lt border}
\gpsetdashtype{gp dt solid}
\gpsetlinewidth{1.00}
\draw[gp path] (1.872,0.985)--(2.052,0.985);
\draw[gp path] (13.447,0.985)--(13.267,0.985);
\node[gp node right] at (1.688,0.985) {0};
\draw[gp path] (1.872,1.612)--(2.052,1.612);
\draw[gp path] (13.447,1.612)--(13.267,1.612);
\node[gp node right] at (1.688,1.612) {2,000};
\draw[gp path] (1.872,2.240)--(2.052,2.240);
\draw[gp path] (13.447,2.240)--(13.267,2.240);
\node[gp node right] at (1.688,2.240) {4,000};
\draw[gp path] (1.872,2.867)--(2.052,2.867);
\draw[gp path] (13.447,2.867)--(13.267,2.867);
\node[gp node right] at (1.688,2.867) {6,000};
\draw[gp path] (1.872,3.495)--(2.052,3.495);
\draw[gp path] (13.447,3.495)--(13.267,3.495);
\node[gp node right] at (1.688,3.495) {8,000};
\draw[gp path] (1.872,4.122)--(2.052,4.122);
\draw[gp path] (13.447,4.122)--(13.267,4.122);
\node[gp node right] at (1.688,4.122) {10,000};
\draw[gp path] (1.872,4.750)--(2.052,4.750);
\draw[gp path] (13.447,4.750)--(13.267,4.750);
\node[gp node right] at (1.688,4.750) {12,000};
\draw[gp path] (1.872,5.377)--(2.052,5.377);
\draw[gp path] (13.447,5.377)--(13.267,5.377);
\node[gp node right] at (1.688,5.377) {14,000};
\draw[gp path] (1.872,0.985)--(1.872,1.165);
\draw[gp path] (1.872,5.691)--(1.872,5.511);
\node[gp node center] at (1.872,0.677) {1};
\draw[gp path] (3.526,0.985)--(3.526,1.165);
\draw[gp path] (3.526,5.691)--(3.526,5.511);
\node[gp node center] at (3.526,0.677) {2};
\draw[gp path] (5.179,0.985)--(5.179,1.165);
\draw[gp path] (5.179,5.691)--(5.179,5.511);
\node[gp node center] at (5.179,0.677) {4};
\draw[gp path] (6.833,0.985)--(6.833,1.165);
\draw[gp path] (6.833,5.691)--(6.833,5.511);
\node[gp node center] at (6.833,0.677) {8};
\draw[gp path] (8.486,0.985)--(8.486,1.165);
\draw[gp path] (8.486,5.691)--(8.486,5.511);
\node[gp node center] at (8.486,0.677) {16};
\draw[gp path] (10.140,0.985)--(10.140,1.165);
\draw[gp path] (10.140,5.691)--(10.140,5.511);
\node[gp node center] at (10.140,0.677) {32};
\draw[gp path] (11.793,0.985)--(11.793,1.165);
\draw[gp path] (11.793,5.691)--(11.793,5.511);
\node[gp node center] at (11.793,0.677) {64};
\draw[gp path] (13.447,0.985)--(13.447,1.165);
\draw[gp path] (13.447,5.691)--(13.447,5.511);
\node[gp node center] at (13.447,0.677) {128};
\draw[gp path] (1.872,5.691)--(1.872,0.985)--(13.447,0.985)--(13.447,5.691)--cycle;
\node[gp node center,rotate=-270] at (0.292,3.338) {Decoding time (seconds)};
\node[gp node center] at (7.659,0.215) {Batch size (sentences)};
\node[gp node right] at (12.531,5.280) {AR Big};
\gpcolor{rgb color={0.200,0.200,0.200}}
\gpsetlinewidth{2.50}
\draw[gp path] (12.715,5.280)--(13.079,5.280);
\draw[gp path] (6.396,5.691)--(6.833,5.251)--(8.486,4.419)--(10.140,4.010)--(11.793,3.857)%
  --(13.447,3.837);
\gpsetpointsize{4.00}
\gppoint{gp mark 1}{(6.833,5.251)}
\gppoint{gp mark 1}{(8.486,4.419)}
\gppoint{gp mark 1}{(10.140,4.010)}
\gppoint{gp mark 1}{(11.793,3.857)}
\gppoint{gp mark 1}{(13.447,3.837)}
\gppoint{gp mark 1}{(12.897,5.280)}
\gpcolor{color=gp lt color border}
\node[gp node right] at (12.531,4.818) {AR Base};
\gpcolor{rgb color={0.800,0.800,0.800}}
\draw[gp path] (12.715,4.818)--(13.079,4.818);
\draw[gp path] (1.872,4.296)--(3.526,3.111)--(5.179,2.447)--(6.833,2.135)--(8.486,1.913)%
  --(10.140,1.814)--(11.793,1.789)--(13.447,1.797);
\gppoint{gp mark 2}{(1.872,4.296)}
\gppoint{gp mark 2}{(3.526,3.111)}
\gppoint{gp mark 2}{(5.179,2.447)}
\gppoint{gp mark 2}{(6.833,2.135)}
\gppoint{gp mark 2}{(8.486,1.913)}
\gppoint{gp mark 2}{(10.140,1.814)}
\gppoint{gp mark 2}{(11.793,1.789)}
\gppoint{gp mark 2}{(13.447,1.797)}
\gppoint{gp mark 2}{(12.897,4.818)}
\gpcolor{color=gp lt color border}
\node[gp node right] at (12.531,4.356) {Large};
\gpcolor{rgb color={0.800,0.000,0.000}}
\draw[gp path] (12.715,4.356)--(13.079,4.356);
\draw[gp path] (1.872,5.000)--(3.526,4.082)--(5.179,3.572)--(6.833,3.352)--(8.486,3.317)%
  --(10.140,3.341)--(11.793,3.382);
\gppoint{gp mark 3}{(1.872,5.000)}
\gppoint{gp mark 3}{(3.526,4.082)}
\gppoint{gp mark 3}{(5.179,3.572)}
\gppoint{gp mark 3}{(6.833,3.352)}
\gppoint{gp mark 3}{(8.486,3.317)}
\gppoint{gp mark 3}{(10.140,3.341)}
\gppoint{gp mark 3}{(11.793,3.382)}
\gppoint{gp mark 3}{(12.897,4.356)}
\gpcolor{color=gp lt color border}
\node[gp node right] at (12.531,3.894) {Base};
\gpcolor{rgb color={1.000,0.800,0.000}}
\draw[gp path] (12.715,3.894)--(13.079,3.894);
\draw[gp path] (1.872,1.785)--(3.526,1.706)--(5.179,1.695)--(6.833,1.708)--(8.486,1.738)%
  --(10.140,1.770)--(11.793,1.804)--(13.447,1.834);
\gppoint{gp mark 4}{(1.872,1.785)}
\gppoint{gp mark 4}{(3.526,1.706)}
\gppoint{gp mark 4}{(5.179,1.695)}
\gppoint{gp mark 4}{(6.833,1.708)}
\gppoint{gp mark 4}{(8.486,1.738)}
\gppoint{gp mark 4}{(10.140,1.770)}
\gppoint{gp mark 4}{(11.793,1.804)}
\gppoint{gp mark 4}{(13.447,1.834)}
\gppoint{gp mark 4}{(12.897,3.894)}
\gpcolor{color=gp lt color border}
\node[gp node right] at (12.531,3.432) {Small};
\gpcolor{rgb color={0.800,0.000,1.000}}
\draw[gp path] (12.715,3.432)--(13.079,3.432);
\draw[gp path] (1.872,1.407)--(3.526,1.376)--(5.179,1.377)--(6.833,1.395)--(8.486,1.424)%
  --(10.140,1.455)--(11.793,1.489)--(13.447,1.523);
\gppoint{gp mark 5}{(1.872,1.407)}
\gppoint{gp mark 5}{(3.526,1.376)}
\gppoint{gp mark 5}{(5.179,1.377)}
\gppoint{gp mark 5}{(6.833,1.395)}
\gppoint{gp mark 5}{(8.486,1.424)}
\gppoint{gp mark 5}{(10.140,1.455)}
\gppoint{gp mark 5}{(11.793,1.489)}
\gppoint{gp mark 5}{(13.447,1.523)}
\gppoint{gp mark 5}{(12.897,3.432)}
\gpcolor{color=gp lt color border}
\node[gp node right] at (12.531,2.970) {Micro};
\gpcolor{rgb color={0.000,0.000,0.800}}
\draw[gp path] (12.715,2.970)--(13.079,2.970);
\draw[gp path] (1.872,1.286)--(3.526,1.266)--(5.179,1.271)--(6.833,1.291)--(8.486,1.318)%
  --(10.140,1.350)--(11.793,1.384)--(13.447,1.418);
\gppoint{gp mark 6}{(1.872,1.286)}
\gppoint{gp mark 6}{(3.526,1.266)}
\gppoint{gp mark 6}{(5.179,1.271)}
\gppoint{gp mark 6}{(6.833,1.291)}
\gppoint{gp mark 6}{(8.486,1.318)}
\gppoint{gp mark 6}{(10.140,1.350)}
\gppoint{gp mark 6}{(11.793,1.384)}
\gppoint{gp mark 6}{(13.447,1.418)}
\gppoint{gp mark 6}{(12.897,2.970)}
\gpcolor{color=gp lt color border}
\node[gp node right] at (12.531,2.508) {Tiny};
\gpcolor{rgb color={0.000,0.800,0.000}}
\draw[gp path] (12.715,2.508)--(13.079,2.508);
\draw[gp path] (1.872,1.062)--(3.526,1.062)--(5.179,1.071)--(6.833,1.084)--(8.486,1.102)%
  --(10.140,1.122)--(11.793,1.144)--(13.447,1.167);
\gppoint{gp mark 7}{(1.872,1.062)}
\gppoint{gp mark 7}{(3.526,1.062)}
\gppoint{gp mark 7}{(5.179,1.071)}
\gppoint{gp mark 7}{(6.833,1.084)}
\gppoint{gp mark 7}{(8.486,1.102)}
\gppoint{gp mark 7}{(10.140,1.122)}
\gppoint{gp mark 7}{(11.793,1.144)}
\gppoint{gp mark 7}{(13.447,1.167)}
\gppoint{gp mark 7}{(12.897,2.508)}
\gpcolor{color=gp lt color border}
\gpsetlinewidth{1.00}
\draw[gp path] (1.872,5.691)--(1.872,0.985)--(13.447,0.985)--(13.447,5.691)--cycle;
%% coordinates of the plot area
\gpdefrectangularnode{gp plot 1}{\pgfpoint{1.872cm}{0.985cm}}{\pgfpoint{13.447cm}{5.691cm}}
\end{tikzpicture}
%% gnuplot variables
}
    \caption{The decoding times to translate the efficiency task test set using various batch size settings, computed on 36 CPU cores.}
    \label{fig:time-cpu}
\end{figure*}
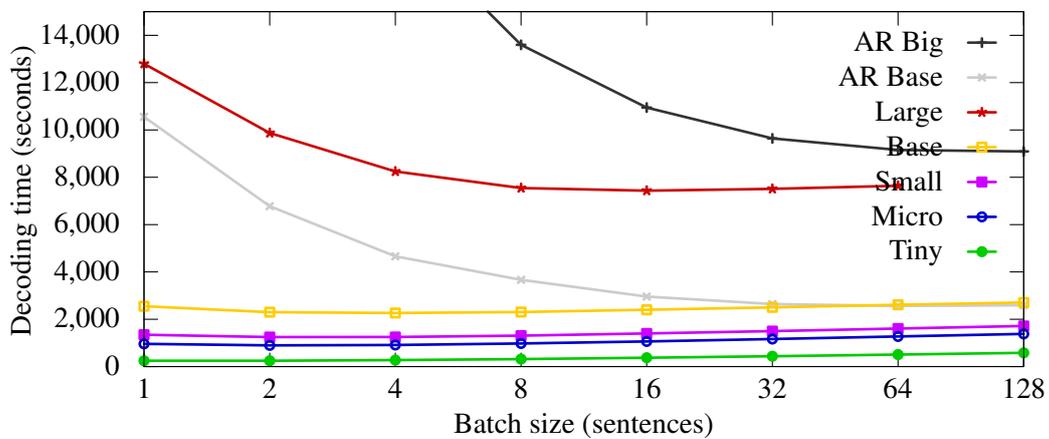

We measure the overall wall time to translate the shared task dataset with
different batching settings on both the GPU and the 36-core CPU. The decoding
times are shown in Figures \ref{fig:time-gpu} and \ref{fig:time-cpu} for the
GPU and CPU times, respectively.  We do not report the single-core CPU
latencies as the decoding speed of the NAR models falls far behind the
efficient AR models in this setup and the translation of the dataset takes too
long.
%\LB{Here the NAR models are actually faster for reasonable batch sizes}

\begin{figure*}
    \centering
    \scalebox{1.0}{\begin{tikzpicture}[gnuplot]
%% generated with GNUPLOT 5.2p8 (Lua 5.3; terminal rev. Nov 2018, script rev. 108)
%% Thu 16 Dec 2021 11:23:09 AM CET
\path (0.000,0.000) rectangle (14.000,6.000);
\gpcolor{color=gp lt color border}
\gpsetlinetype{gp lt border}
\gpsetdashtype{gp dt solid}
\gpsetlinewidth{1.00}
\draw[gp path] (1.872,0.985)--(2.052,0.985);
\draw[gp path] (13.447,0.985)--(13.267,0.985);
\node[gp node right] at (1.688,0.985) {0};
\draw[gp path] (1.872,1.769)--(2.052,1.769);
\draw[gp path] (13.447,1.769)--(13.267,1.769);
\node[gp node right] at (1.688,1.769) {5,000};
\draw[gp path] (1.872,2.554)--(2.052,2.554);
\draw[gp path] (13.447,2.554)--(13.267,2.554);
\node[gp node right] at (1.688,2.554) {10,000};
\draw[gp path] (1.872,3.338)--(2.052,3.338);
\draw[gp path] (13.447,3.338)--(13.267,3.338);
\node[gp node right] at (1.688,3.338) {15,000};
\draw[gp path] (1.872,4.122)--(2.052,4.122);
\draw[gp path] (13.447,4.122)--(13.267,4.122);
\node[gp node right] at (1.688,4.122) {20,000};
\draw[gp path] (1.872,4.907)--(2.052,4.907);
\draw[gp path] (13.447,4.907)--(13.267,4.907);
\node[gp node right] at (1.688,4.907) {25,000};
\draw[gp path] (1.872,5.691)--(2.052,5.691);
\draw[gp path] (13.447,5.691)--(13.267,5.691);
\node[gp node right] at (1.688,5.691) {30,000};
\draw[gp path] (1.872,0.985)--(1.872,1.165);
\draw[gp path] (1.872,5.691)--(1.872,5.511);
\node[gp node center] at (1.872,0.677) {1};
\draw[gp path] (3.526,0.985)--(3.526,1.165);
\draw[gp path] (3.526,5.691)--(3.526,5.511);
\node[gp node center] at (3.526,0.677) {2};
\draw[gp path] (5.179,0.985)--(5.179,1.165);
\draw[gp path] (5.179,5.691)--(5.179,5.511);
\node[gp node center] at (5.179,0.677) {4};
\draw[gp path] (6.833,0.985)--(6.833,1.165);
\draw[gp path] (6.833,5.691)--(6.833,5.511);
\node[gp node center] at (6.833,0.677) {8};
\draw[gp path] (8.486,0.985)--(8.486,1.165);
\draw[gp path] (8.486,5.691)--(8.486,5.511);
\node[gp node center] at (8.486,0.677) {16};
\draw[gp path] (10.140,0.985)--(10.140,1.165);
\draw[gp path] (10.140,5.691)--(10.140,5.511);
\node[gp node center] at (10.140,0.677) {32};
\draw[gp path] (11.793,0.985)--(11.793,1.165);
\draw[gp path] (11.793,5.691)--(11.793,5.511);
\node[gp node center] at (11.793,0.677) {64};
\draw[gp path] (13.447,0.985)--(13.447,1.165);
\draw[gp path] (13.447,5.691)--(13.447,5.511);
\node[gp node center] at (13.447,0.677) {128};
\draw[gp path] (1.872,5.691)--(1.872,0.985)--(13.447,0.985)--(13.447,5.691)--cycle;
\node[gp node center,rotate=-270] at (0.016,3.338) {Decoding time (seconds)};
\node[gp node center] at (7.659,0.215) {Batch size (sentences)};
\node[gp node right] at (12.531,5.280) {AR Large};
\gpcolor{rgb color={0.200,0.200,0.200}}
\gpsetlinewidth{2.50}
\draw[gp path] (12.715,5.280)--(13.079,5.280);
\draw[gp path] (6.498,5.691)--(6.833,5.100)--(8.486,3.611)--(10.140,2.752)--(11.793,2.385)%
  --(13.447,2.153);
\gpsetpointsize{4.00}
\gppoint{gp mark 1}{(6.833,5.100)}
\gppoint{gp mark 1}{(8.486,3.611)}
\gppoint{gp mark 1}{(10.140,2.752)}
\gppoint{gp mark 1}{(11.793,2.385)}
\gppoint{gp mark 1}{(13.447,2.153)}
\gppoint{gp mark 1}{(12.897,5.280)}
\gpcolor{color=gp lt color border}
\node[gp node right] at (12.531,4.818) {AR Base};
\gpcolor{rgb color={0.800,0.800,0.800}}
\draw[gp path] (12.715,4.818)--(13.079,4.818);
\draw[gp path] (5.602,5.691)--(6.833,3.888)--(8.486,2.724)--(10.140,2.032)--(11.793,1.691)%
  --(13.447,1.520);
\gppoint{gp mark 2}{(6.833,3.888)}
\gppoint{gp mark 2}{(8.486,2.724)}
\gppoint{gp mark 2}{(10.140,2.032)}
\gppoint{gp mark 2}{(11.793,1.691)}
\gppoint{gp mark 2}{(13.447,1.520)}
\gppoint{gp mark 2}{(12.897,4.818)}
\gpcolor{color=gp lt color border}
\node[gp node right] at (12.531,4.356) {Large};
\gpcolor{rgb color={0.800,0.000,0.000}}
\draw[gp path] (12.715,4.356)--(13.079,4.356);
\draw[gp path] (1.872,3.206)--(3.526,2.598)--(5.179,2.252)--(6.833,2.084)--(8.486,1.997)%
  --(10.140,1.935)--(11.793,1.890)--(13.447,1.860);
\gppoint{gp mark 3}{(1.872,3.206)}
\gppoint{gp mark 3}{(3.526,2.598)}
\gppoint{gp mark 3}{(5.179,2.252)}
\gppoint{gp mark 3}{(6.833,2.084)}
\gppoint{gp mark 3}{(8.486,1.997)}
\gppoint{gp mark 3}{(10.140,1.935)}
\gppoint{gp mark 3}{(11.793,1.890)}
\gppoint{gp mark 3}{(13.447,1.860)}
\gppoint{gp mark 3}{(12.897,4.356)}
\gpcolor{color=gp lt color border}
\node[gp node right] at (12.531,3.894) {Base};
\gpcolor{rgb color={1.000,0.800,0.000}}
\draw[gp path] (12.715,3.894)--(13.079,3.894);
\draw[gp path] (1.872,2.353)--(3.526,1.853)--(5.179,1.593)--(6.833,1.479)--(8.486,1.419)%
  --(10.140,1.383)--(11.793,1.361)--(13.447,1.346);
\gppoint{gp mark 4}{(1.872,2.353)}
\gppoint{gp mark 4}{(3.526,1.853)}
\gppoint{gp mark 4}{(5.179,1.593)}
\gppoint{gp mark 4}{(6.833,1.479)}
\gppoint{gp mark 4}{(8.486,1.419)}
\gppoint{gp mark 4}{(10.140,1.383)}
\gppoint{gp mark 4}{(11.793,1.361)}
\gppoint{gp mark 4}{(13.447,1.346)}
\gppoint{gp mark 4}{(12.897,3.894)}
\gpcolor{color=gp lt color border}
\node[gp node right] at (12.531,3.432) {Small};
\gpcolor{rgb color={0.800,0.000,1.000}}
\draw[gp path] (12.715,3.432)--(13.079,3.432);
\draw[gp path] (1.872,1.772)--(3.526,1.498)--(5.179,1.358)--(6.833,1.295)--(8.486,1.259)%
  --(10.140,1.241)--(11.793,1.230)--(13.447,1.222);
\gppoint{gp mark 5}{(1.872,1.772)}
\gppoint{gp mark 5}{(3.526,1.498)}
\gppoint{gp mark 5}{(5.179,1.358)}
\gppoint{gp mark 5}{(6.833,1.295)}
\gppoint{gp mark 5}{(8.486,1.259)}
\gppoint{gp mark 5}{(10.140,1.241)}
\gppoint{gp mark 5}{(11.793,1.230)}
\gppoint{gp mark 5}{(13.447,1.222)}
\gppoint{gp mark 5}{(12.897,3.432)}
\gpcolor{color=gp lt color border}
\node[gp node right] at (12.531,2.970) {Micro};
\gpcolor{rgb color={0.000,0.000,0.800}}
\draw[gp path] (12.715,2.970)--(13.079,2.970);
\draw[gp path] (1.872,1.578)--(3.526,1.371)--(5.179,1.278)--(6.833,1.233)--(8.486,1.207)%
  --(10.140,1.196)--(11.793,1.186)--(13.447,1.181);
\gppoint{gp mark 6}{(1.872,1.578)}
\gppoint{gp mark 6}{(3.526,1.371)}
\gppoint{gp mark 6}{(5.179,1.278)}
\gppoint{gp mark 6}{(6.833,1.233)}
\gppoint{gp mark 6}{(8.486,1.207)}
\gppoint{gp mark 6}{(10.140,1.196)}
\gppoint{gp mark 6}{(11.793,1.186)}
\gppoint{gp mark 6}{(13.447,1.181)}
\gppoint{gp mark 6}{(12.897,2.970)}
\gpcolor{color=gp lt color border}
\node[gp node right] at (12.531,2.508) {Tiny};
\gpcolor{rgb color={0.000,0.800,0.000}}
\draw[gp path] (12.715,2.508)--(13.079,2.508);
\draw[gp path] (1.872,1.286)--(3.526,1.188)--(5.179,1.136)--(6.833,1.111)--(8.486,1.100)%
  --(10.140,1.093)--(11.793,1.089)--(13.447,1.087);
\gppoint{gp mark 7}{(1.872,1.286)}
\gppoint{gp mark 7}{(3.526,1.188)}
\gppoint{gp mark 7}{(5.179,1.136)}
\gppoint{gp mark 7}{(6.833,1.111)}
\gppoint{gp mark 7}{(8.486,1.100)}
\gppoint{gp mark 7}{(10.140,1.093)}
\gppoint{gp mark 7}{(11.793,1.089)}
\gppoint{gp mark 7}{(13.447,1.087)}
\gppoint{gp mark 7}{(12.897,2.508)}
\gpcolor{color=gp lt color border}
\gpsetlinewidth{1.00}
\draw[gp path] (1.872,5.691)--(1.872,0.985)--(13.447,0.985)--(13.447,5.691)--cycle;
%% coordinates of the plot area
\gpdefrectangularnode{gp plot 1}{\pgfpoint{1.872cm}{0.985cm}}{\pgfpoint{13.447cm}{5.691cm}}
\end{tikzpicture}
%% gnuplot variables
}
    \caption{The decoding times to translate the efficiency task test set using various batch size settings, computed on a single Nvidia Pascal P100 GPU.}
    \label{fig:time-p100}
\end{figure*}
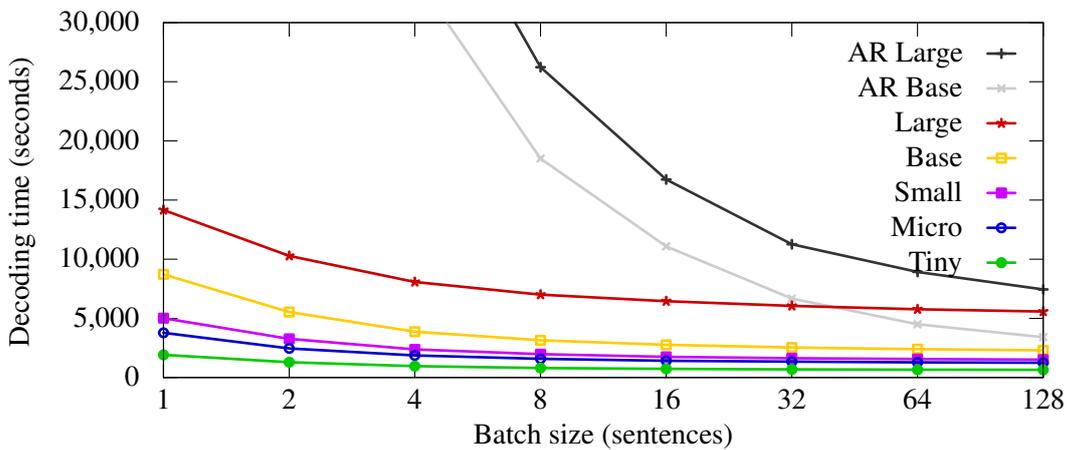

We can see that in case of GPU decoding that all models benefit from having
larger batch sizes. However, the non-autoregressive models are much faster when
the batch size is small. We also ran the evaluation on an Nvidia Pascal P100
GPU, which showed that when the batch size is large enough, autoregressive
models eventually match the speed of non-autoregressive
models. We show the decoding times on the Pascal GPU in Figure \ref{fig:time-p100}. In Table \ref{tab:p100-latency}, we compare the latencies measured on the Pascal GPU to some of the related NAR approaches that report results on this GPU type. 
Due to implementation reasons, the maximum batch size for our NAR models is around 220 sentences. 
%\LB{You need to explain why you do not show the result in the figure because the figure shows that NAR is faster. Please be even clearer here that it is due to NAR not fitting in memory with higher batch sizes, and even put the AR result in the figure itself if you can}

%\JH{discuss what happens for larger batch sizes}

\begin{table}
    \centering
    \begin{tabular}{lr}
        \toprule
        Model & Latency (ms)\\
        \midrule
        \citet{gu2017nonautoregressive} & 39 \\
        \citet{wang2019non} & 22 \\
        \citet{sun2019fast} & 37 \\
        \midrule
        Ours -- Large &         14 \\
        \bottomrule
    \end{tabular}
    \caption{The comparison of the decoding time of various NAR models for a single sentence in a batch on a P100 GPU. Note that this table should serve merely as an illustration, since the results were measured on different datasets.}
    %\LB{Need to reference and discuss in the text and also clarify what you mean by latency in caption. Nice result though}
    \label{tab:p100-latency}
\end{table}

\begin{table*}[ht!]
  \centering

  \begin{tabular}{lrrrrrr}
    \toprule
    & \multicolumn{3}{c}{Translation quality } & \multicolumn{3}{c}{Decoding time (seconds)} \\
    & {\small ChrF} & {\small COMET} & {\small BLEU} & {\small GPU, b>1} & {\small GPU, b=1} & {\small CPU, b>1} \\
    \midrule
    Edinburgh base \citep{Behnke-wmt21-speed} & 61.5 & 0.527 & 55.3 & 140 & 16,851 & 500 \\
    \midrule
    AR -- Large (teacher) & 59.2 & 0.411 & 50.5 & 1,918 & {\it > 24h} & 9,090 \\
    AR -- Base (student) & 59.5 & 0.455 & 51.6 & 1,465 & {\it > 24h} & 2,587 \\
    \addlinespace
    NAR -- Large & 58.6 & 0.149 & 47.8 & 782 & 7,020 & 7,434 \\
    NAR -- Micro & 57.3 & -0.008 & 43.5 & 311 & 2,322 & 897 \\
    \bottomrule
  \end{tabular}

  \caption{A comparison of our AR and NAR models with one of the submissions to the WMT~21 efficiency task. We show the results of automatic translation quality evaluation using three different metrics, and the decoding time to translate the test set using a GPU and 36-core CPU with either latency (b=1) or batched (b>1) decoding.}
  \label{tab:comparison}
\end{table*}

\paragraph{Comparison with Efficient AR Models.}

%\LB{Is this also on the million sentence test set?}\JH{yes}
In Table \ref{tab:comparison}, we present a comparison on the million sentence test set with ``Edinburgh base'', one of the leading submissions in the WMT~21
efficiency task \citep{Behnke-wmt21-speed}, which uses the deep encoder -- shallow decoder architecture \citep{kasai2021deep}.
First, we see that using three different evaluation metrics (ChrF, COMET, and BLEU), our models lag behind the Edinburgh base model.
In line with our previous observation, we see a considerable drop in the COMET score values.
In terms of decoding speed, the only scenario in which the non-autoregressive model is better is on GPU with batch size 1. This is in line
with our intuition that the parallelization potential brought by the GPU is utilized more efficiently by the NAR model. On one hand,
larger batches open up the parallelization possibilities to AR models. On the other hand, limited parallelization potential (in form of CPU decoding) reduces the differences between AR and NAR models. The batch size of the Edinburgh base model was 1,280 in the batched decoding setup.  

% -----------------------------------------------------------------------------
\section{Conclusions}%
\label{sec:conclusions}
% -----------------------------------------------------------------------------

In this paper, we challenge the evaluation methodology adopted by the
research on non-autoregressive models for NMT.

We argue that in terms of translation quality, the evaluation should include newer test sets and metrics other 
than BLEU (particularly COMET and ChrF). This will provide more insight and put the results into the context of the recent research. 

From the decoding speed perspective, we should bear in mind various use-cases for the model deployment, 
such as the hardware environment or batching conditions.  Preferably, the research should evaluate the speed gains across a range of scenarios. Finally, given that the latency condition -- translation of one sentence at a time on a GPU -- already translates too fast to be perceived by human users of MT, there is currently no compelling scenario that warrants the deployment of NAR models. 

\section*{Acknowledgments}
\label{SEC:ACK}

This project received funding from the European Union’s Horizon 2020 research and innovation programmes under grant agreements 825299 and 825303 (GoURMET, Bergamot), and from the Czech Science Foundation grant 19-26934X (NEUREM3) of the Czech Science Foundation.
%The project was further supported by the UK Engineering and Physical Sciences Research Council (EPSRC) fellowship grant EP/S001271/1 (MTStretch). 
Our work has been using data provided by the LINDAT/CLARIAH-CZ Research Infrastructure, supported by the Ministry of Education, Youth and Sports of the Czech Republic (Project No. LM2018101).

\bibliography{anthology,custom}

\begin{thebibliography}{40}
\expandafter\ifx\csname natexlab\endcsname\relax\def\natexlab#1{#1}\fi

\bibitem[{Akhbardeh et~al.(2021)Akhbardeh, Arkhangorodsky, Biesialska, Bojar,
  Chatterjee, Chaudhary, Costa-jussa, España-Bonet, Fan, Federmann, Freitag,
  Graham, Grundkiewicz, Haddow, Harter, Heafield, Homan, Huck,
  Amponsah-Kaakyire, Kasai, Khashabi, Knight, Kocmi, Koehn, Lourie, Monz,
  Morishita, Nagata, Nagesh, Nakazawa, Negri, Pal, Tapo, Turchi, Vydrin, and
  Zampieri}]{akhbardeh-etal-2021-findings}
Farhad Akhbardeh, Arkady Arkhangorodsky, Magdalena Biesialska, Ondřej Bojar,
  Rajen Chatterjee, Vishrav Chaudhary, Marta~R. Costa-jussa, Cristina
  España-Bonet, Angela Fan, Christian Federmann, Markus Freitag, Yvette
  Graham, Roman Grundkiewicz, Barry Haddow, Leonie Harter, Kenneth Heafield,
  Christopher Homan, Matthias Huck, Kwabena Amponsah-Kaakyire, Jungo Kasai,
  Daniel Khashabi, Kevin Knight, Tom Kocmi, Philipp Koehn, Nicholas Lourie,
  Christof Monz, Makoto Morishita, Masaaki Nagata, Ajay Nagesh, Toshiaki
  Nakazawa, Matteo Negri, Santanu Pal, Allahsera~Auguste Tapo, Marco Turchi,
  Valentin Vydrin, and Marcos Zampieri. 2021.
\newblock \href {https://aclanthology.org/2021.wmt-1.1} {Findings of the 2021
  conference on machine translation ({WMT21})}.
\newblock In \emph{Proceedings of the Sixth Conference on Machine Translation},
  pages 1--93, Online. Association for Computational Linguistics.

\bibitem[{Amodei et~al.(2016)Amodei, Ananthanarayanan, Anubhai, Bai,
  Battenberg, Case, Casper, Catanzaro, Cheng, Chen, Chen, Chen, Chen,
  Chrzanowski, Coates, Diamos, Ding, Du, Elsen, Engel, Fang, Fan, Fougner, Gao,
  Gong, Hannun, Han, Johannes, Jiang, Ju, Jun, LeGresley, Lin, Liu, Liu, Li,
  Li, Ma, Narang, Ng, Ozair, Peng, Prenger, Qian, Quan, Raiman, Rao, Satheesh,
  Seetapun, Sengupta, Srinet, Sriram, Tang, Tang, Wang, Wang, Wang, Wang, Wang,
  Wang, Wu, Wei, Xiao, Xie, Xie, Yogatama, Yuan, Zhan, and
  Zhu}]{amodei2016deep}
Dario Amodei, Sundaram Ananthanarayanan, Rishita Anubhai, Jingliang Bai, Eric
  Battenberg, Carl Case, Jared Casper, Bryan Catanzaro, Qiang Cheng, Guoliang
  Chen, Jie Chen, Jingdong Chen, Zhijie Chen, Mike Chrzanowski, Adam Coates,
  Greg Diamos, Ke~Ding, Niandong Du, Erich Elsen, Jesse Engel, Weiwei Fang,
  Linxi Fan, Christopher Fougner, Liang Gao, Caixia Gong, Awni Hannun, Tony
  Han, Lappi Johannes, Bing Jiang, Cai Ju, Billy Jun, Patrick LeGresley, Libby
  Lin, Junjie Liu, Yang Liu, Weigao Li, Xiangang Li, Dongpeng Ma, Sharan
  Narang, Andrew Ng, Sherjil Ozair, Yiping Peng, Ryan Prenger, Sheng Qian,
  Zongfeng Quan, Jonathan Raiman, Vinay Rao, Sanjeev Satheesh, David Seetapun,
  Shubho Sengupta, Kavya Srinet, Anuroop Sriram, Haiyuan Tang, Liliang Tang,
  Chong Wang, Jidong Wang, Kaifu Wang, Yi~Wang, Zhijian Wang, Zhiqian Wang,
  Shuang Wu, Likai Wei, Bo~Xiao, Wen Xie, Yan Xie, Dani Yogatama, Bin Yuan, Jun
  Zhan, and Zhenyao Zhu. 2016.
\newblock \href {https://proceedings.mlr.press/v48/amodei16.html} {Deep speech
  2 : End-to-end speech recognition in english and mandarin}.
\newblock In \emph{Proceedings of The 33rd International Conference on Machine
  Learning}, volume~48 of \emph{Proceedings of Machine Learning Research},
  pages 173--182, New York, New York, USA. PMLR.

\bibitem[{Bahdanau et~al.(2016)Bahdanau, Cho, and Bengio}]{bahdanau2016neural}
Dzmitry Bahdanau, Kyunghyun Cho, and Yoshua Bengio. 2016.
\newblock \href {http://arxiv.org/abs/1409.0473} {Neural machine translation by
  jointly learning to align and translate}.

\bibitem[{Ba{\~n}{\'o}n et~al.(2020)Ba{\~n}{\'o}n, Chen, Haddow, Heafield,
  Hoang, Espl{\`a}-Gomis, Forcada, Kamran, Kirefu, Koehn, Ortiz~Rojas,
  Pla~Sempere, Ram{\'\i}rez-S{\'a}nchez, Sarr{\'\i}as, Strelec, Thompson,
  Waites, Wiggins, and Zaragoza}]{banon-etal-2020-paracrawl}
Marta Ba{\~n}{\'o}n, Pinzhen Chen, Barry Haddow, Kenneth Heafield, Hieu Hoang,
  Miquel Espl{\`a}-Gomis, Mikel~L. Forcada, Amir Kamran, Faheem Kirefu, Philipp
  Koehn, Sergio Ortiz~Rojas, Leopoldo Pla~Sempere, Gema
  Ram{\'\i}rez-S{\'a}nchez, Elsa Sarr{\'\i}as, Marek Strelec, Brian Thompson,
  William Waites, Dion Wiggins, and Jaume Zaragoza. 2020.
\newblock \href {https://doi.org/10.18653/v1/2020.acl-main.417} {{P}ara{C}rawl:
  Web-scale acquisition of parallel corpora}.
\newblock In \emph{Proceedings of the 58th Annual Meeting of the Association
  for Computational Linguistics}, pages 4555--4567, Online. Association for
  Computational Linguistics.

\bibitem[{Behnke et~al.(2021)Behnke, Bogoychev, Aji, Heafield, Nail, Zhu,
  Tchistiakova, van~der Linde, Chen, Kashyap, and
  Grundkiewicz}]{Behnke-wmt21-speed}
Maximiliana Behnke, Nikolay Bogoychev, Alham~Fikri Aji, Kenneth Heafield,
  Graeme Nail, Qianqian Zhu, Svetlana Tchistiakova, Jelmer van~der Linde,
  Pinzhen Chen, Sidharth Kashyap, and Roman Grundkiewicz. 2021.
\newblock \href {https://kheafield.com/papers/edinburgh/wmt21-speed.pdf}
  {Efficient machine translation with model pruning and quantization}.
\newblock In \emph{Proceedings of the Conference on Machine Translation at the
  2021 Conference on Empirical Methods in Natural Language Processing}, Punta
  Cana, Dominican Republic.

\bibitem[{Bojar et~al.(2014)Bojar, Buck, Federmann, Haddow, Koehn, Leveling,
  Monz, Pecina, Post, Saint-Amand, Soricut, Specia, and
  Tamchyna}]{bojar-etal-2014-findings}
Ond{\v{r}}ej Bojar, Christian Buck, Christian Federmann, Barry Haddow, Philipp
  Koehn, Johannes Leveling, Christof Monz, Pavel Pecina, Matt Post, Herve
  Saint-Amand, Radu Soricut, Lucia Specia, and Ale{\v{s}} Tamchyna. 2014.
\newblock \href {https://doi.org/10.3115/v1/W14-3302} {Findings of the 2014
  workshop on statistical machine translation}.
\newblock In \emph{Proceedings of the Ninth Workshop on Statistical Machine
  Translation}, pages 12--58, Baltimore, Maryland, USA. Association for
  Computational Linguistics.

\bibitem[{Bojar et~al.(2016)Bojar, Chatterjee, Federmann, Graham, Haddow, Huck,
  Jimeno~Yepes, Koehn, Logacheva, Monz, Negri, N{\'e}v{\'e}ol, Neves, Popel,
  Post, Rubino, Scarton, Specia, Turchi, Verspoor, and
  Zampieri}]{bojar-etal-2016-findings}
Ond{\v{r}}ej Bojar, Rajen Chatterjee, Christian Federmann, Yvette Graham, Barry
  Haddow, Matthias Huck, Antonio Jimeno~Yepes, Philipp Koehn, Varvara
  Logacheva, Christof Monz, Matteo Negri, Aur{\'e}lie N{\'e}v{\'e}ol, Mariana
  Neves, Martin Popel, Matt Post, Raphael Rubino, Carolina Scarton, Lucia
  Specia, Marco Turchi, Karin Verspoor, and Marcos Zampieri. 2016.
\newblock \href {https://doi.org/10.18653/v1/W16-2301} {Findings of the 2016
  conference on machine translation}.
\newblock In \emph{Proceedings of the First Conference on Machine Translation:
  Volume 2, Shared Task Papers}, pages 131--198, Berlin, Germany. Association
  for Computational Linguistics.

\bibitem[{Caswell et~al.(2019)Caswell, Chelba, and
  Grangier}]{caswell-etal-2019-tagged}
Isaac Caswell, Ciprian Chelba, and David Grangier. 2019.
\newblock \href {https://doi.org/10.18653/v1/W19-5206} {Tagged
  back-translation}.
\newblock In \emph{Proceedings of the Fourth Conference on Machine Translation
  (Volume 1: Research Papers)}, pages 53--63, Florence, Italy. Association for
  Computational Linguistics.

\bibitem[{Chen et~al.(2021)Chen, Helcl, Germann, Burchell, Bogoychev,
  Miceli~Barone, Waldendorf, Birch, and Heafield}]{chen-wmt21-news}
Pinzhen Chen, Jind\v{r}ich Helcl, Ulrich Germann, Laurie Burchell, Nikolay
  Bogoychev, Antonio~Valerio Miceli~Barone, Jonas Waldendorf, Alexandra Birch,
  and Kenneth Heafield. 2021.
\newblock \href {https://kheafield.com/papers/edinburgh/wmt21-news.pdf} {The
  {U}niversity of {Edinburgh's} {E}nglish-{G}erman and {E}nglish-{H}ausa
  submissions to the {WMT21} news translation task}.
\newblock In \emph{Proceedings of the Conference on Machine Translation at the
  2021 Conference on Empirical Methods in Natural Language Processing}, Punta
  Cana, Dominican Republic.

\bibitem[{Ghazvininejad et~al.(2020)Ghazvininejad, Karpukhin, Zettlemoyer, and
  Levy}]{ghazvininejad2020aligned}
Marjan Ghazvininejad, Vladimir Karpukhin, Luke Zettlemoyer, and Omer Levy.
  2020.
\newblock \href {http://proceedings.mlr.press/v119/ghazvininejad20a.html}
  {Aligned cross entropy for non-autoregressive machine translation}.
\newblock In \emph{Proceedings of the 37th International Conference on Machine
  Learning}, volume 119 of \emph{Proceedings of Machine Learning Research},
  pages 3515--3523. PMLR.

\bibitem[{Ghazvininejad et~al.(2019)Ghazvininejad, Levy, Liu, and
  Zettlemoyer}]{ghazvininejad-etal-2019-mask}
Marjan Ghazvininejad, Omer Levy, Yinhan Liu, and Luke Zettlemoyer. 2019.
\newblock \href {https://doi.org/10.18653/v1/D19-1633} {Mask-predict: Parallel
  decoding of conditional masked language models}.
\newblock In \emph{Proceedings of the 2019 Conference on Empirical Methods in
  Natural Language Processing and the 9th International Joint Conference on
  Natural Language Processing (EMNLP-IJCNLP)}, pages 6112--6121, Hong Kong,
  China. Association for Computational Linguistics.

\bibitem[{Graves et~al.(2006)Graves, Fern{\'a}ndez, Gomez, and
  Schmidhuber}]{graves2006connectionist}
Alex Graves, Santiago Fern{\'a}ndez, Faustino Gomez, and J\"{u}rgen
  Schmidhuber. 2006.
\newblock Connectionist temporal classification: Labelling unsegmented sequence
  data with recurrent neural networks.
\newblock In \emph{Proceedings of the 23rd International Conference on Machine
  Learning}, pages 369--376, Pittsburgh, PA, USA. JMLR.org.

\bibitem[{Gu et~al.(2018)Gu, Bradbury, Xiong, Li, and
  Socher}]{gu2017nonautoregressive}
Jiatao Gu, James Bradbury, Caiming Xiong, Victor O.~K. Li, and Richard Socher.
  2018.
\newblock \href {https://openreview.net/forum?id=B1l8BtlCb} {Non-autoregressive
  neural machine translation}.
\newblock In \emph{6th International Conference on Learning Representations,
  {ICLR} 2018}, Vancouver, BC, Canada.

\bibitem[{Gu and Kong(2021)}]{gu-kong-2021-fully}
Jiatao Gu and Xiang Kong. 2021.
\newblock \href {https://doi.org/10.18653/v1/2021.findings-acl.11} {Fully
  non-autoregressive neural machine translation: Tricks of the trade}.
\newblock In \emph{Findings of the Association for Computational Linguistics:
  ACL-IJCNLP 2021}, pages 120--133, Online. Association for Computational
  Linguistics.

\bibitem[{Heafield et~al.(2021)Heafield, Zhu, and
  Grundkiewicz}]{heafield-etal-2021-findings}
Kenneth Heafield, Qianqian Zhu, and Roman Grundkiewicz. 2021.
\newblock \href {https://kheafield.com/papers/edinburgh/wmt21-speedtask.pdf}
  {Findings of the {WMT} 2021 shared task on efficient translation}.
\newblock In \emph{Proceedings of the Conference on Machine Translation at the
  2021 Conference on Empirical Methods in Natural Language Processing}, Punta
  Cana, Dominican Republic.

\bibitem[{Junczys-Dowmunt(2018)}]{junczys-dowmunt-2018-dual}
Marcin Junczys-Dowmunt. 2018.
\newblock \href {https://doi.org/10.18653/v1/W18-6478} {Dual conditional
  cross-entropy filtering of noisy parallel corpora}.
\newblock In \emph{Proceedings of the Third Conference on Machine Translation:
  Shared Task Papers}, pages 888--895, Belgium, Brussels. Association for
  Computational Linguistics.

\bibitem[{Junczys-Dowmunt et~al.(2018)Junczys-Dowmunt, Grundkiewicz, Dwojak,
  Hoang, Heafield, Neckermann, Seide, Germann, Aji, Bogoychev, Martins, and
  Birch}]{junczys-dowmunt-etal-2018-marian}
Marcin Junczys-Dowmunt, Roman Grundkiewicz, Tomasz Dwojak, Hieu Hoang, Kenneth
  Heafield, Tom Neckermann, Frank Seide, Ulrich Germann, Alham~Fikri Aji,
  Nikolay Bogoychev, Andr{\'e} F.~T. Martins, and Alexandra Birch. 2018.
\newblock \href {https://doi.org/10.18653/v1/P18-4020} {{M}arian: Fast neural
  machine translation in {C}++}.
\newblock In \emph{Proceedings of {ACL} 2018, System Demonstrations}, pages
  116--121, Melbourne, Australia. Association for Computational Linguistics.

\bibitem[{Kaiser et~al.(2018)Kaiser, Bengio, Roy, Vaswani, Parmar, Uszkoreit,
  and Shazeer}]{kaiser2018fast}
Lukasz Kaiser, Samy Bengio, Aurko Roy, Ashish Vaswani, Niki Parmar, Jakob
  Uszkoreit, and Noam Shazeer. 2018.
\newblock Fast decoding in sequence models using discrete latent variables.
\newblock In \emph{International Conference on Machine Learning}, pages
  2390--2399. PMLR.

\bibitem[{Kasai et~al.(2021)Kasai, Pappas, Peng, Cross, and
  Smith}]{kasai2021deep}
Jungo Kasai, Nikolaos Pappas, Hao Peng, James Cross, and Noah Smith. 2021.
\newblock \href {https://openreview.net/forum?id=KpfasTaLUpq} {Deep encoder,
  shallow decoder: Reevaluating non-autoregressive machine translation}.
\newblock In \emph{International Conference on Learning Representations}.

\bibitem[{Kim and Rush(2016)}]{kim-rush-2016-sequence}
Yoon Kim and Alexander~M. Rush. 2016.
\newblock \href {https://doi.org/10.18653/v1/D16-1139} {Sequence-level
  knowledge distillation}.
\newblock In \emph{Proceedings of the 2016 Conference on Empirical Methods in
  Natural Language Processing}, pages 1317--1327, Austin, Texas. Association
  for Computational Linguistics.

\bibitem[{Kingma and Ba(2014)}]{kingma2014adam}
Diederik~P Kingma and Jimmy Ba. 2014.
\newblock Adam: {A} method for stochastic optimization.
\newblock \emph{CoRR}, abs/1412.6980.

\bibitem[{Kocmi et~al.(2021)Kocmi, Federmann, Grundkiewicz, Junczys-Dowmunt,
  Matsushita, and Menezes}]{kocmi2021ship}
Tom Kocmi, Christian Federmann, Roman Grundkiewicz, Marcin Junczys-Dowmunt,
  Hitokazu Matsushita, and Arul Menezes. 2021.
\newblock To ship or not to ship: An extensive evaluation of automatic metrics
  for machine translation.
\newblock \emph{arXiv preprint arXiv:2107.10821}.

\bibitem[{Koehn(2005)}]{koehn-2005-europarl}
Philipp Koehn. 2005.
\newblock \href {https://aclanthology.org/2005.mtsummit-papers.11} {{E}uroparl:
  A parallel corpus for statistical machine translation}.
\newblock In \emph{Proceedings of Machine Translation Summit X: Papers}, pages
  79--86, Phuket, Thailand.

\bibitem[{Lee et~al.(2018)Lee, Mansimov, and Cho}]{lee-etal-2018-deterministic}
Jason Lee, Elman Mansimov, and Kyunghyun Cho. 2018.
\newblock \href {https://doi.org/10.18653/v1/D18-1149} {Deterministic
  non-autoregressive neural sequence modeling by iterative refinement}.
\newblock In \emph{Proceedings of the 2018 Conference on Empirical Methods in
  Natural Language Processing}, pages 1173--1182, Brussels, Belgium.
  Association for Computational Linguistics.

\bibitem[{Libovick{\'y} and Helcl(2018)}]{libovicky-helcl-2018-end}
Jind{\v{r}}ich Libovick{\'y} and Jind{\v{r}}ich Helcl. 2018.
\newblock \href {https://doi.org/10.18653/v1/D18-1336} {End-to-end
  non-autoregressive neural machine translation with connectionist temporal
  classification}.
\newblock In \emph{Proceedings of the 2018 Conference on Empirical Methods in
  Natural Language Processing}, pages 3016--3021, Brussels, Belgium.
  Association for Computational Linguistics.

\bibitem[{Mathur et~al.(2020)Mathur, Baldwin, and
  Cohn}]{mathur-etal-2020-tangled}
Nitika Mathur, Timothy Baldwin, and Trevor Cohn. 2020.
\newblock \href {https://doi.org/10.18653/v1/2020.acl-main.448} {Tangled up in
  {BLEU}: Reevaluating the evaluation of automatic machine translation
  evaluation metrics}.
\newblock In \emph{Proceedings of the 58th Annual Meeting of the Association
  for Computational Linguistics}, pages 4984--4997, Online. Association for
  Computational Linguistics.

\bibitem[{Papineni et~al.(2002)Papineni, Roukos, Ward, and
  Zhu}]{papineni-etal-2002-bleu}
Kishore Papineni, Salim Roukos, Todd Ward, and Wei-Jing Zhu. 2002.
\newblock \href {https://doi.org/10.3115/1073083.1073135} {{B}leu: a method for
  automatic evaluation of machine translation}.
\newblock In \emph{Proceedings of the 40th Annual Meeting of the Association
  for Computational Linguistics}, pages 311--318, Philadelphia, Pennsylvania,
  USA. Association for Computational Linguistics.

\bibitem[{Popovi{\'c}(2015)}]{popovic-2015-chrf}
Maja Popovi{\'c}. 2015.
\newblock \href {https://doi.org/10.18653/v1/W15-3049} {chr{F}: character
  n-gram {F}-score for automatic {MT} evaluation}.
\newblock In \emph{Proceedings of the Tenth Workshop on Statistical Machine
  Translation}, pages 392--395, Lisbon, Portugal. Association for Computational
  Linguistics.

\bibitem[{Post(2018)}]{post-2018-call}
Matt Post. 2018.
\newblock \href {https://doi.org/10.18653/v1/W18-6319} {A call for clarity in
  reporting {BLEU} scores}.
\newblock In \emph{Proceedings of the Third Conference on Machine Translation:
  Research Papers}, pages 186--191, Brussels, Belgium. Association for
  Computational Linguistics.

\bibitem[{Qian et~al.(2021)Qian, Zhou, Bao, Wang, Qiu, Zhang, Yu, and
  Li}]{qian-etal-2021-glancing}
Lihua Qian, Hao Zhou, Yu~Bao, Mingxuan Wang, Lin Qiu, Weinan Zhang, Yong Yu,
  and Lei Li. 2021.
\newblock \href {https://doi.org/10.18653/v1/2021.acl-long.155} {Glancing
  transformer for non-autoregressive neural machine translation}.
\newblock In \emph{Proceedings of the 59th Annual Meeting of the Association
  for Computational Linguistics and the 11th International Joint Conference on
  Natural Language Processing (Volume 1: Long Papers)}, pages 1993--2003,
  Online. Association for Computational Linguistics.

\bibitem[{Rei et~al.(2020)Rei, Stewart, Farinha, and
  Lavie}]{rei-etal-2020-comet}
Ricardo Rei, Craig Stewart, Ana~C Farinha, and Alon Lavie. 2020.
\newblock \href {https://doi.org/10.18653/v1/2020.emnlp-main.213} {{COMET}: A
  neural framework for {MT} evaluation}.
\newblock In \emph{Proceedings of the 2020 Conference on Empirical Methods in
  Natural Language Processing (EMNLP)}, pages 2685--2702, Online. Association
  for Computational Linguistics.

\bibitem[{Rozis and Skadi{\c{n}}{\v{s}}(2017)}]{rozis-skadins-2017-tilde}
Roberts Rozis and Raivis Skadi{\c{n}}{\v{s}}. 2017.
\newblock \href {https://aclanthology.org/W17-0235} {Tilde {MODEL} -
  multilingual open data for {EU} languages}.
\newblock In \emph{Proceedings of the 21st Nordic Conference on Computational
  Linguistics}, pages 263--265, Gothenburg, Sweden. Association for
  Computational Linguistics.

\bibitem[{Saharia et~al.(2020)Saharia, Chan, Saxena, and
  Norouzi}]{saharia-etal-2020-non}
Chitwan Saharia, William Chan, Saurabh Saxena, and Mohammad Norouzi. 2020.
\newblock \href {https://doi.org/10.18653/v1/2020.emnlp-main.83}
  {Non-autoregressive machine translation with latent alignments}.
\newblock In \emph{Proceedings of the 2020 Conference on Empirical Methods in
  Natural Language Processing (EMNLP)}, pages 1098--1108, Online. Association
  for Computational Linguistics.

\bibitem[{Schwenk et~al.(2019)Schwenk, Chaudhary, Sun, Gong, and
  Guzm{\'a}n}]{schwenk2019wikimatrix}
Holger Schwenk, Vishrav Chaudhary, Shuo Sun, Hongyu Gong, and Francisco
  Guzm{\'a}n. 2019.
\newblock Wikimatrix: Mining 135{M} parallel sentences in 1620 language pairs
  from wikipedia.
\newblock \emph{arXiv preprint arXiv:1907.05791}.

\bibitem[{Sennrich et~al.(2016)Sennrich, Haddow, and
  Birch}]{sennrich-etal-2016-improving}
Rico Sennrich, Barry Haddow, and Alexandra Birch. 2016.
\newblock \href {https://doi.org/10.18653/v1/P16-1009} {Improving neural
  machine translation models with monolingual data}.
\newblock In \emph{Proceedings of the 54th Annual Meeting of the Association
  for Computational Linguistics (Volume 1: Long Papers)}, pages 86--96, Berlin,
  Germany. Association for Computational Linguistics.

\bibitem[{Sun et~al.(2019)Sun, Li, Wang, He, Lin, and Deng}]{sun2019fast}
Zhiqing Sun, Zhuohan Li, Haoqing Wang, Di~He, Zi~Lin, and Zhihong Deng. 2019.
\newblock \href
  {https://proceedings.neurips.cc/paper/2019/file/74563ba21a90da13dacf2a73e3ddefa7-Paper.pdf}
  {Fast structured decoding for sequence models}.
\newblock In \emph{Advances in Neural Information Processing Systems},
  volume~32, pages 3016--3026. Curran Associates, Inc.

\bibitem[{Tiedemann(2012)}]{tiedemann-2012-parallel}
J{\"o}rg Tiedemann. 2012.
\newblock \href
  {http://www.lrec-conf.org/proceedings/lrec2012/pdf/463_Paper.pdf} {Parallel
  data, tools and interfaces in {OPUS}}.
\newblock In \emph{Proceedings of the Eighth International Conference on
  Language Resources and Evaluation ({LREC}'12)}, pages 2214--2218, Istanbul,
  Turkey. European Language Resources Association (ELRA).

\bibitem[{Vaswani et~al.(2017)Vaswani, Shazeer, Parmar, Uszkoreit, Jones,
  Gomez, Kaiser, and Polosukhin}]{vaswani-etal-2017-attention}
Ashish Vaswani, Noam Shazeer, Niki Parmar, Jakob Uszkoreit, Llion Jones,
  Aidan~N Gomez, \L~ukasz Kaiser, and Illia Polosukhin. 2017.
\newblock \href
  {https://proceedings.neurips.cc/paper/2017/file/3f5ee243547dee91fbd053c1c4a845aa-Paper.pdf}
  {Attention is all you need}.
\newblock In \emph{Advances in Neural Information Processing Systems},
  volume~30. Curran Associates, Inc.

\bibitem[{Wang et~al.(2019)Wang, Tian, He, Qin, Zhai, and Liu}]{wang2019non}
Yiren Wang, Fei Tian, Di~He, Tao Qin, ChengXiang Zhai, and Tie-Yan Liu. 2019.
\newblock Non-autoregressive machine translation with auxiliary regularization.
\newblock In \emph{Proceedings of the AAAI Conference on Artificial
  Intelligence}, volume~33, pages 5377--5384.

\bibitem[{Zhou et~al.(2020)Zhou, Gu, and Neubig}]{zhou-etal-2020-understanding}
Chunting Zhou, Jiatao Gu, and Graham Neubig. 2020.
\newblock \href {https://openreview.net/forum?id=BygFVAEKDH} {Understanding
  knowledge distillation in non-autoregressive machine translation}.
\newblock In \emph{8th International Conference on Learning Representations,
  {ICLR} 2020, Addis Ababa, Ethiopia, April 26-30, 2020}. OpenReview.net.

\end{thebibliography}
\bibliographystyle{acl_natbib}

\end{document}